\documentclass[a4paper,fleqn]{cas-dc}

\usepackage[authoryear]{natbib}
\usepackage{subfig}
\usepackage{placeins}
\usepackage{float}
\usepackage{hyperref}
\usepackage{flafter}
\usepackage{graphicx}

\def\tsc#1{\csdef{#1}{\textsc{\lowercase{#1}}
\xspace}}

\tsc{WGM}
\tsc{QE}
\tsc{EP}
\tsc{PMS}
\tsc{BEC}
\tsc{DE}

\begin{document}
\let\WriteBookmarks\relax
\def\floatpagepagefraction{1}
\def\textpagefraction{.001}

\shorttitle{Probabilistic road classification in historical maps
using synthetic data and deep learning}

\shortauthors{Mühlematter et~al.}

\title [mode = title]{Probabilistic road classification in historical maps
using synthetic data and deep learning}                      


%
\author[1]{Dominik J. Mühlematter}[orcid=0000-0001-6800-9114]
\cormark[1]
\fnmark[1]

\ead{dmuehlema@ethz.ch}

\author[1]{Sebastian Schweizer}
\fnmark[1]

\author[1]{Chenjing Jiao}

\author[1]{Xue Xia}

\author[1,2]{Magnus Heitzler}

\author[1]{Lorenz Hurni}


\fntext[1]{Equal contribution.}



\affiliation[1]{organization={Institute of Cartography and Geoinformation},
    addressline={ETH Zürich}, 
    city={Zürich},
    country={Switzerland}}

\affiliation[2]{organization={Heitzler Geoinformatik},
    country={Germany}}

\cortext[cor1]{Corresponding author}


\begin{abstract}
Historical maps are invaluable for analyzing long-term changes in transportation and spatial development, offering a rich source of data for evolutionary studies. However, digitizing and classifying road networks from these maps is often prohibitively expensive and time-consuming, limiting their widespread use. Recent advancements in deep learning have made automatic road extraction from historical maps feasible, yet these methods typically require large amounts of expensive labeled training data. To address this challenge, we introduce a novel framework that integrates deep learning with geoinformation, computer-based painting, and image processing methodologies. This framework enables the extraction and classification of roads from historical maps using only road geometries without needing road class labels for training. The process begins with cascaded training of a binary segmentation model to extract road geometries, followed by morphological operations, skeletonization, vectorization, and filtering algorithms. Synthetic training data is then generated by a painting function that artificially re-paints road segments using predefined symbology for road classes. Using this synthetic data, a deep ensemble is trained to generate pixel-wise probabilities for road classes to mitigate distribution shift. These predictions are then discretized along the extracted road geometries. Subsequently, further processing is employed to classify entire roads, enabling the identification of potential changes in road classes and resulting in a labeled road class dataset.  
Our method achieved completeness and correctness scores of over 94\% and 92\%, respectively, for road class 2, the most prevalent class in the two Siegfried Map sheets from Switzerland used for testing. This research offers a powerful tool for urban planning and transportation decision-making by efficiently extracting and classifying roads from historical maps, and potentially even satellite images.
\end{abstract}




\maketitle

\section{Introduction}
\label{sec:intro}

Historical maps are invaluable for examining geographic features from past eras, often serving as the sole source of professionally surveyed data before the advent of aerial imagery \citep{chiang2020, Acvi2022Extraction}. Preserving and digitizing these maps not only protects valuable historical cartographic information but also enhances our ability to analyze and understand geographic and anthropogenic changes over time \citep{uhl2022, jacobson1940history}. The digital documentation of infrastructure, such as extracted road geometries from historical maps, is crucial for informed decision-making in transportation, significantly impacting regional development and society \citep{casali2019topological, zhao2015statistical}. Beyond road geometries, other semantic features like road class information offer valuable insights into historical logistics and military operations \citep{Ekim2021}. Moreover, there is growing interest in the use of historical maps for spatial data conflation \citep{chen2008,tong2014} and for expanding public map databases  \citep{Swisstopo2024, arcanummaps2024}.

The application of long-term road data analysis is limited by the costly and time-consuming process of vectorization, especially for more extended temporal map series and larger areas. The task becomes even more laborious when additional semantic information, such as road classifications, is required. Therefore, extensive research has focused on automatically extracting road data from raster maps. Various approaches leverage the parallel characteristics of roads for feature extraction. Early attempts successfully detected parallel road lines in scanned maps \citep{watanabe2001, dhar2006extraction}. Improved versions, such as those by \cite{chiang2009automatic} and \cite{chiang2013}, can distinguish single lines from double lines, though they still struggle with dashed lines. Other popular approaches utilize clustering algorithms for Color Image  Segmentation (CIS) \citep{cheng1995mean} applied to road extraction from historical maps \citep{Jiao2021methods}. These methods often require integration with other techniques such as morphological operations \citep{kasturi1988information} and line tracing to enhance performance \citep{chiang2009automatic, chiang2013, dhar2006extraction}.

Recently, research covering road extraction from historical maps has been dominated by computer vision algorithms using neural networks. Promising results were generated by using variants of U-Net architectures \citep{U_Net, jiao2022fast, jiao2024novel}, also combined with self-attention layers \citep{Vaswani2017, Acvi2022Extraction}. Further, there is a trend toward transfer learning by finetuning pre-trained models for extracting roads in historical maps to increase performance in settings with limited training data \citep{Ekim2021, Acvi2022Extraction}.

Through the limited availability of training data, research has investigated the creation and use of synthetic training data to increase the amount of training data artificially. \cite{jiao2022aug} showed that road segmentation performance can be improved with advanced data augmentation techniques by applying random transformations only to road features while leaving other map symbology, such as text, coordinate grids, or settlements, untransformed. \cite{jiao2022fast} used symbol reconstruction to create synthetic training data.

While synthetic data can be used to increase the amount of training data, the distribution shift between synthetic and real data often poses a problem for training models that can generalize well \citep{Zhang2021DeepSL}. This issue arises from insufficient knowledge about specific areas within the input space through the lack of training data, also referred to as epistemic uncertainty. Therefore, for a model to be resilient to different input distributions than those used in training, it is crucial to ensure accurate predictive uncertainty \citep{Ovadia2019Shift, uncertaintyquantificationimagebasedtraffic}. Much research has investigated approximate Bayesian inference through variational inference \citep{Graves2011PracticalNetworks, Blundell2015WeightNetworks, Neal1996BayesianNetworks, Chen2014StochasticCarlo, Welling2011BayesianDynamics}, since the analytical computation of the posterior in neural networks is intractable. However, these methods often fail to accurately capture high-dimensional data \citep{Gustafsson2019EvaluatingVision}. Promising approaches for calibrating neural networks include variants of deep ensembles \citep{Lakshminarayanan2017SimpleEnsembles, Havasi2020TrainingPrediction, Turkoglu2022, LoRA_ensemble, Gal2015Dropout}.

While most research focuses on extracting road geometries alone, some studies have also conducted road classification. However, these approaches usually require expensive road class labeled training data \citep{Ekim2021, can2021automatic}. Recently, \cite{jiao2024novel} introduced a method that leverages deep learning to extract road geometries, followed by symbol painting for template matching-based road classification, without needing road class labeled training data. Inspired by this idea, we developed a novel approach for road vectorization and classification, utilizing deep learning-based road classification without the need for class labels in the training data.

First, we apply a \textit{cascaded training} approach to a binary segmentation model for extracting road geometries from historical maps. This involves sequentially pre-training the model on larger datasets before fine-tuning it on the historical map data. Morphological operations, filtering, and vectorization follow this. Then, symbol painting is used to create synthetic training data with road class labels by randomly overpainting roads with specific class symbology in the training data. Subsequently, a deep ensemble is employed to predict pixel-wise class probabilities \citep{Lakshminarayanan2017SimpleEnsembles}, which are then combined with the previously extracted road geometries. Zonal statistics within a buffer around each road are calculated by averaging the predicted class probabilities. Subsequently, we analyze the predicted probabilities along each road segment to identify locations where the road class shifts. This enables precise road class categorization even within a single extracted road segment between two intersections.

The approach results in a vectorized road dataset evaluated on the Swiss \textit{Siegfried Map}. Our method is the first to employ synthetic data for training a neural network to perform road classification without the need for labeled training data. Our study's promising results demonstrate our approach's effectiveness and its potential for future applications. We published our code on GitHub\footnote{\url{https://github.com/DominikM198/ProbRoadClass-DeepLearning}}. The weights for applying our method to the \textit{Siegfried Map} and for transfer learning applications are available on Hugging Face\footnote{\url{https://huggingface.co/DominikM198/ProbRoadClass-DeepLearning}}.

\section{Data}
\label{sec:data}
\begin{figure}[h]
    \centering
    \captionsetup[subfloat]{labelformat=empty}
    \subfloat[Class 1]{
        \includegraphics[width=0.14\linewidth]{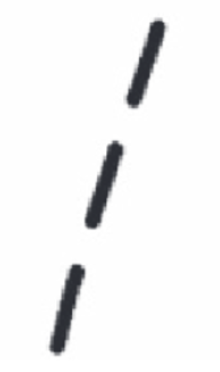} \label{fig:road_class1}
    }%
    \hspace*{0.25cm}
    \subfloat[Class 2]{
        \includegraphics[width=0.14\linewidth]{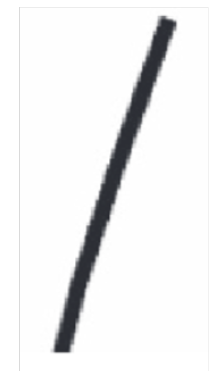} \label{fig:road_class2}
    }%
    \hspace*{0.25cm}
    \subfloat[Class 3]{
        \includegraphics[width=0.14\linewidth]{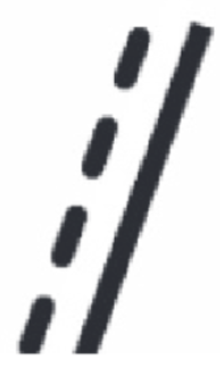} \label{fig:road_class3}
    }%
    \hspace*{0.25cm}
    \subfloat[Class 4]{
        \includegraphics[width=0.14\linewidth]{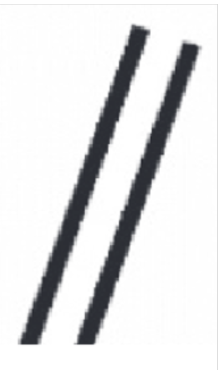} \label{fig:road_class4}
    }%
    \hspace*{0.25cm}
    \subfloat[Class 5]{
        \includegraphics[width=0.14\linewidth]{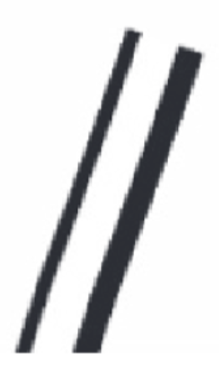} \label{fig:road_class5}
    }%
    \caption{The road class symbols of \textit{Siegfried Map}. Adapted from \cite{jiao2024novel}.}
    \label{fig:road_classes}
\end{figure}

\begin{figure}[h!]
    \centering
    \subfloat[\textit{Siegfried Map}]{{\includegraphics[width=0.42\linewidth]{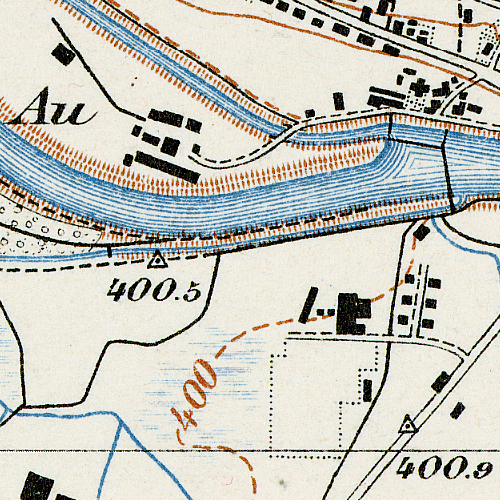}\label{fig:siegfried_example} }}%
    \qquad
    \subfloat[\textit{Swiss Map}]{{\includegraphics[width=0.42\linewidth]{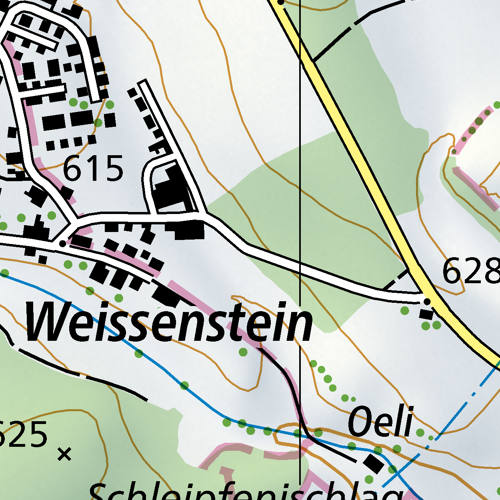}}\label{fig:swissmap_example}  }%
     \caption{Example patches of the two map series used for the study. Geodata © Swisstopo\footref{swisstopo}.}
    \label{fig:shap_values}
\end{figure}

The Swiss \textit{Siegfried Map} series, published between 1872 and 1949, stands as one of the historical map collections of Switzerland \citep{gotsch2002siegfried}. This detailed topographical series illustrates both natural features—such as rivers, moorlands, and forests—and human-made elements, including roads, buildings, railways, and place names. The Swiss Federal Office of Topography (Swisstopo\footnote{\url{https://www.swisstopo.admin.ch/en} \label{swisstopo}}) digitized these maps into raster format. The individual sheets were subsequently georeferenced using the intersection points of the coordinate grid \citep{heitzler2018modular}. A patch of the map is shown in Figure~\ref{fig:siegfried_example}. Each scanned map sheet measures $7000 \times 4800$ pixels. The specific maps discussed in this paper are at a 1:25,000 scale and have been scanned with a spatial resolution of 1.25 meters per pixel. Figure~\ref{fig:road_classes} presents the five road classes present in the \textit{Siegfried Map} \citep{jiao2022fast}:

\begin{itemize}
\item Class 1: Walking path ("Fussweg") 
\item Class 2: Dirt road or mule track ("Feld- oder Saumweg") 
\item Class 3: Driveway without reinforcement ("Fahrweg ohne Kunstanlage") 
\item Class 4: Reinforced road 3–5 meters wide ("Kunststrasse 3-5 Meter Breite") 
\item Class 5: Reinforced road wider than 5 meters ("Kunststrasse über 5 Meter Breite") 
\end{itemize}

The available training data consists of road centerlines in the city of Zurich, originally produced for an internal project by the Institute of Cartography and Geoinformation at ETH Zurich. However, the road classes are not labeled in the training data. For the validation set, we use another map sheet without road class labels, while two map sheets with road class labeled data are used as ground truth for evaluation. 

Given the limited size of the \textit{Siegfried Map} dataset, we pre-trained the model on a larger dataset for the same task to enhance performance. Specifically, we used 19 map sheets from the current Swiss national map provided by Swisstopo\footref{swisstopo} (Figure~\ref{fig:swissmap_example}). Specifically, we used the \textit{Swiss Map Raster~25} for model input and the \textit{Swiss Map Vector~25} for road geometries. In the rest of the paper, we will refer to these as \textit{Swiss Map}. More details about the datasets and the map sheets can be found in Appendix~\ref{Appendix:data}.

\section{Methods}
\label{sec:method}
\begin{figure*}[h!]
    \centering
    \includegraphics[width=1.0\textwidth]{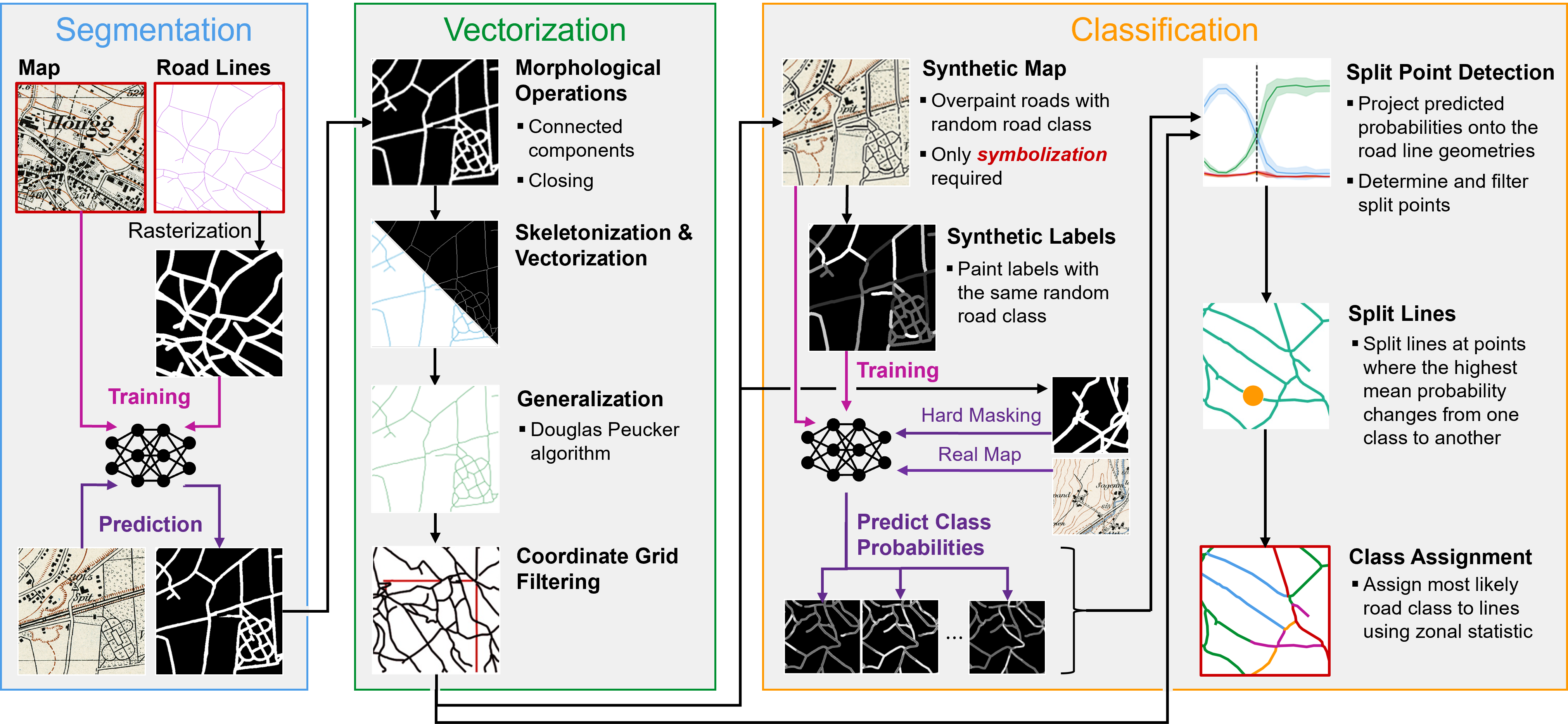}
    \caption{Diagram illustrating the proposed methodology. The input data, including the \textit{Siegfried Map}, unlabeled road centerlines, and the symbol definitions of the five road classes, are shown in red. The black arrows indicate the dependencies on intermediate results from previous parts. Pink arrows represent the neural network training, while violet arrows indicate predictions used for evaluation or application of our method. }Geodata © Swisstopo\footref{swisstopo}.
    \label{fig:schema}
\end{figure*}

 Figure~\ref{fig:schema} shows a schematic representation of the approach pursued. First, we employ a neural network to extract road geometries from historical maps, followed by morphological operations, vectorization, generalization, and filtering. Next, synthetic training data is generated by overpainting roads with class-specific symbology. Deep learning is then used to predict pixel-wise class probabilities, which are integrated with the road geometries. We analyze these probabilities along each road segment to detect class transitions along a route and accurately categorize the roads.

\subsection{Segmentation}
The objective of the segmentation part is to train a neural network to classify pixels as road or non-road. Later, the resulting segmentation output is used to derive road geometries as a vector dataset.

\subsubsection{Pre-processing}\label{sec:seg_preprocessing}
As input for training, validation and testing of the segmentation model, $500~\times~500~$pixel tiles are used, each with $125~$pixel overlap to mitigate boundary effects. Additionally, we rasterize the road centerlines to create binary labels at the same resolution and extent as the \textit{Siegfried Map} sheet. This is done using a uniform line width of $10~$pixels, corresponding to a real-world road width of $12.5~$meters. Applying the same line width for all road categories simplifies training data creation by using only road geometry without additional semantic information. Road geometries for training are available only for the city of Zurich, resulting in partial labels for the four \textit{Siegfried Map} sheets used for training. During training, pixels without ground truth data are ignored; therefore, we rasterize a mask for each tile to indicate the availability of pixel-wise ground truth data. Finally, we have 912 tiles for training, 241 tiles for validation, and 1'160 tiles for testing, each corresponding to a spatial extent of $625~\times~625~$ meters.

\subsubsection{Segmentation model}
\label{sec:segmentation_model}
\begin{figure*}[h]
    \centering
    \includegraphics[width=0.8\textwidth]{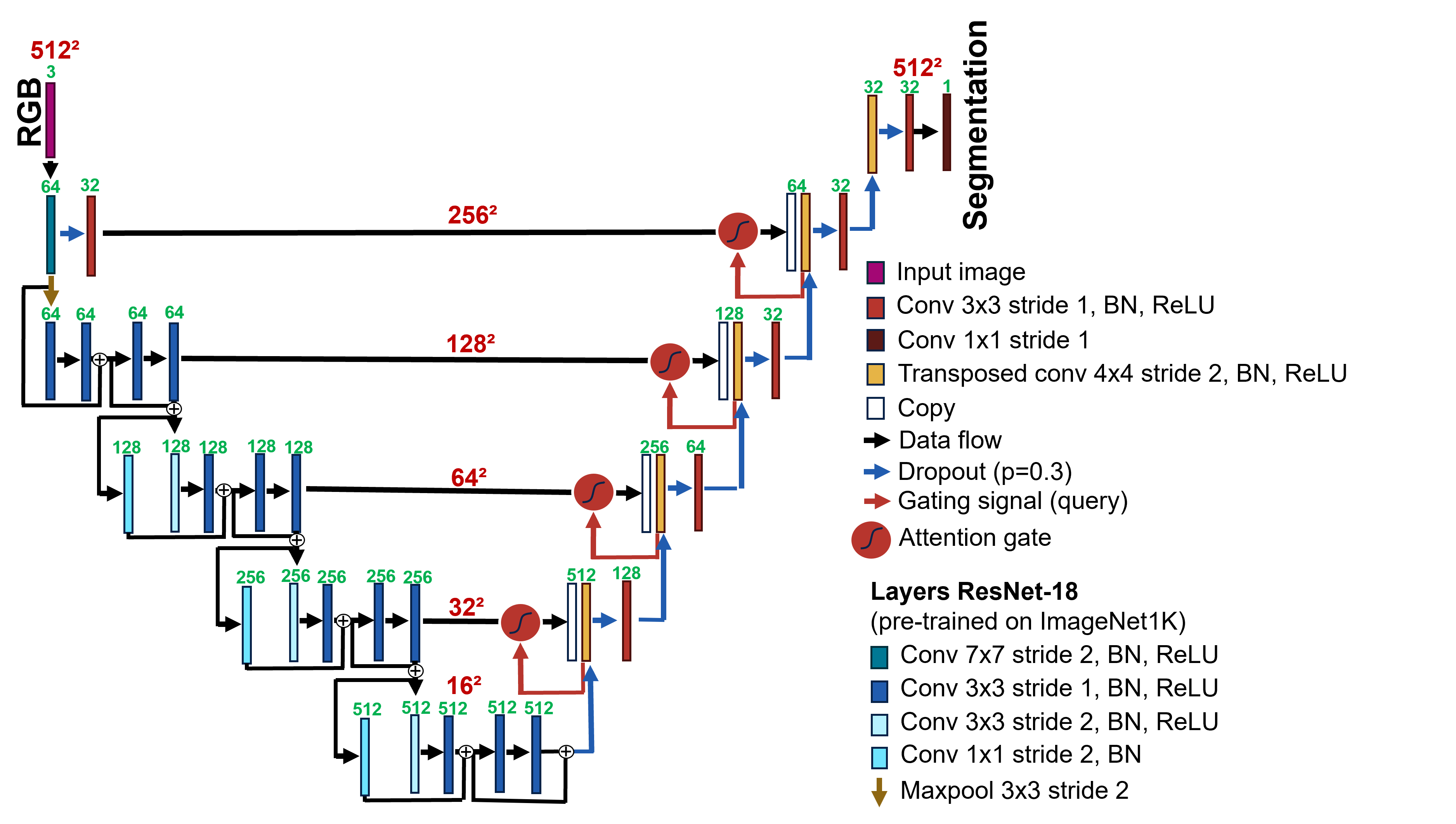}
    \caption{\textit{Attention ResU-Net}: Binary image segmentation architecture with a total of 15'190'373 parameters.
}
    \label{fig:Res-U-Net}
\end{figure*}

We developed a fully convolutional network, \textit{Attention ResU-Net}, which draws inspiration from the U-Net architecture \citep{U_Net}. Our model uses a ResNet-18 pre-trained classification model as the encoder \citep{He2015DeepRL}, as shown in Figure~\ref{fig:Res-U-Net}. We initialize the encoder with weights from \textit{ImageNet} training \citep{Deng2009ImageNet}, capitalizing on their established performance across various transfer learning tasks, such as semantic image segmentation \citep{ImageNetSegmentation}, remote sensing \citep{RemoteSensingImageNet}, and even sound classification \citep{AudioImageNet}. We employed a \textit{cascaded training} approach: In addition to leveraging transfer learning by initializing the encoder with pre-trained weights from the classification model, we afterwards pre-trained the entire \textit{Attention ResU-Net} on the \textit{Swiss Map} for road extraction. Following this, the model weights were fine-tuned for the downstream task on the \textit{Siegfried Map}.

The decoder of our network upsamples the feature map produced by the encoder at multiple resolutions using transposed convolutions \citep{deconvolution}, incorporating dropout for regularization \citep{droput_training}. A batch normalization layer \citep{batchnorm} and a ReLU activation function follow each convolution. We integrated additive attention gates, as proposed by \cite{attention_u_net}, into our segmentation model to focus on target structures with different shapes and sizes. This helps the model to ignore irrelevant areas and emphasize important features for the skip connections, as successfully demonstrated in other research \citep{attention_u_net, attention_u_net_application}. 

We trained the model for 50 epochs using the Adam optimizer \citep{adam} and Dice Loss \citep{DiceLoss}, data augmentation, and early stopping based on the Intersection over Union (IoU) score on the validation set \citep{EarlyStopping}, which was evaluated after each epoch. Additional details regarding training, model selection, hyperparameter tuning,  pre-training, and data augmentation are provided in Appendix~\ref{Appendix:binary_segmentation}.

\subsection{Vectorization}
This Section focuses on post-processing the segmentation results to generate a vectorized road dataset.

\subsubsection{Map stiching}\label{sec:map_stiching}
Map stitching can be seen as the inverse process of tiling. First, the predicted tiles are cropped to $250~\times~250~$pixels to remove the $125$-pixel overlap from the pre-processed tiles. Afterwards, the cropped tiles are stitched together. This results in georeferenced raster predictions of roads for the entire \textit{Siegfried Map} sheets.

\subsubsection{Morphological operations}
The stitched predictions from the segmentation are refined using two morphological operations. First, a connected components analysis (CCA) is performed, which detects connected regions in a binary image, allowing the identification of areas with the same binary label \citep{bailey2007single}. We then eliminate predicted roads with an area of less than $100~$pixels, as empirically proposed by \cite{jiao2024novel}, since small areas are likely noise rather than representing actual roads, which typically have larger interconnected regions.
Secondly, a morphological closing operation is applied. This operation uses a sliding window to perform local adjustments across the image  \citep{bovik2009essential}. The closing operation helps maintain topology after vectorization, ensuring that disconnected predictions are linked. It also improves vectorization by filling holes within road predictions, preventing the creation of dual road axes around such holes during skeletonization. A uniform $3~\times~3$~kernel is used for this operation, implemented with the Python package OpenCV \citep{opencv}.

\subsubsection{Skeletonization and vectorization}
To convert the raster data into vector data, we first perform skeletonization, a morphological operation that shrinks the areas to one-pixel-wide lines representing the road axes. We use the algorithm developed by \cite{skeletonization_lee}, implemented in the Python package Scikit-Image \citep{scikit_image}.

Converting pixel-based lines into vector data poses a challenge, as existing algorithms primarily output points or areas rather than lines. To address this, we developed our own vectorization algorithm. It utilizes the 8-neighbourhood to identify lines between source and possible target pixels. This algorithm generates one vector line per pair of connected pixels. We conducted subsequent dissolve and generalization operations, where small line segments were merged while disjoint features remained separate. This approach ensures a topologically correct dataset where each line represents a road between two intersections.

\subsubsection{Generalization}
Following a multipart to singlepart operation, the dataset underwent generalization using the Douglas-Peucker algorithm \citep{douglas_peucker}, with a distance parameter set to $1.9~\mathrm{m}$. Given the raster resolution of $1.25~\mathrm{m}$ (equivalent to  $1.77~\mathrm{m}$ in the diagonal direction), the chosen distance of  $1.9~\mathrm{m}$ is appropriate for the task. This value was selected based partly on prior research \citep{jiao2024novel} and qualitative analysis using the validation set.

\subsubsection{Coordinate grid filtering}
After pre-training the segmentation model with the \textit{Swiss Map} and fine-tuning it with the \textit{Siegfried Map}, both of which feature a similar coordinate grid representation, the model is largely capable of distinguishing coordinate grid lines from roads. However, there may still be instances where some coordinate grid lines are mistakenly classified as roads, often appearing as short lines that create slight zigzag patterns at intersections of correctly predicted roads. This occurs because the intersection point of skeletonization is based on the center of mass. Hence, we utilize the properties of the coordinate grid for additional enhancements. Since the grid comprises only horizontal or vertical lines, and we have knowledge about the coordinates of the grid, we can filter out the horizontal lines using the following criteria:

\begin{itemize}
    \item The $y$-coordinates of all vertices of a line have to be within a certain buffer around one of the known $y$-coordinates of the coordinate grid.
    \item The sum of all subsequent differences of the $y$-coordinates should be approximately zero.
    \begin{equation}
        \sum_{i=2}^{N_{\text{vertices}}} y_i - y_{i-1} \approx 0.
    \end{equation}
    \item As verification criteria, the summation of differences in $x$-direction should be at least one order of magnitude greater than those in the $y$-direction.
    \begin{equation}
        10 \cdot \left|\sum_{i=2}^{N_{\text{vertices}}} y_i - y_{i-1}\right| < \left|\sum_{i=2}^{N_{\text{vertices}}} x_i - x_{i-1}\right|.
    \end{equation}
\end{itemize}

Vertical lines may be filtered analogously.

\subsection{Classification}
After segmenting and vectorizing the sheets of the \textit{Siegfried Map}, we obtained a vector dataset. However, this dataset lacks road class labels. In this section, we leverage neural networks trained on synthetic data to classify and label the roads in the derived vector dataset.

\subsubsection{Synthetic training and validation data}
\begin{figure}[h]
    \subfloat[Original \textit{Siegfried Map}]{{\includegraphics[width=0.45\linewidth]{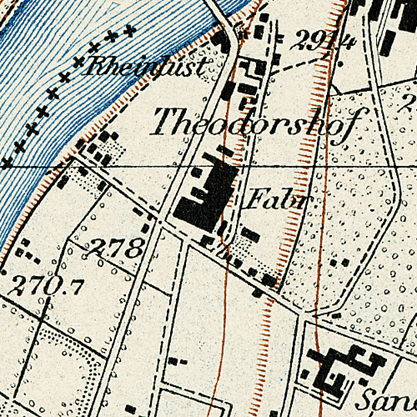}\label{fig:original_siegfried}}}%
    \qquad
    \subfloat[Vectorized predictions]{{\includegraphics[width=0.45\linewidth]{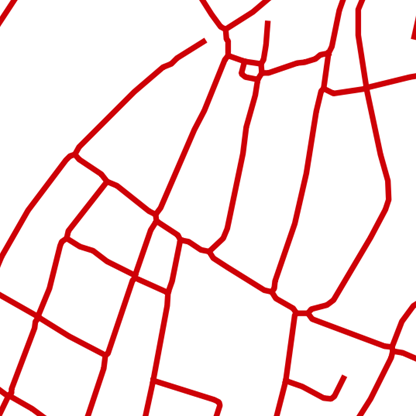} \label{fig:vectorized_predictions}}}%
    \qquad
    \subfloat[Synthetic \textit{Siegfried Map}]{{\includegraphics[width=0.45\linewidth]{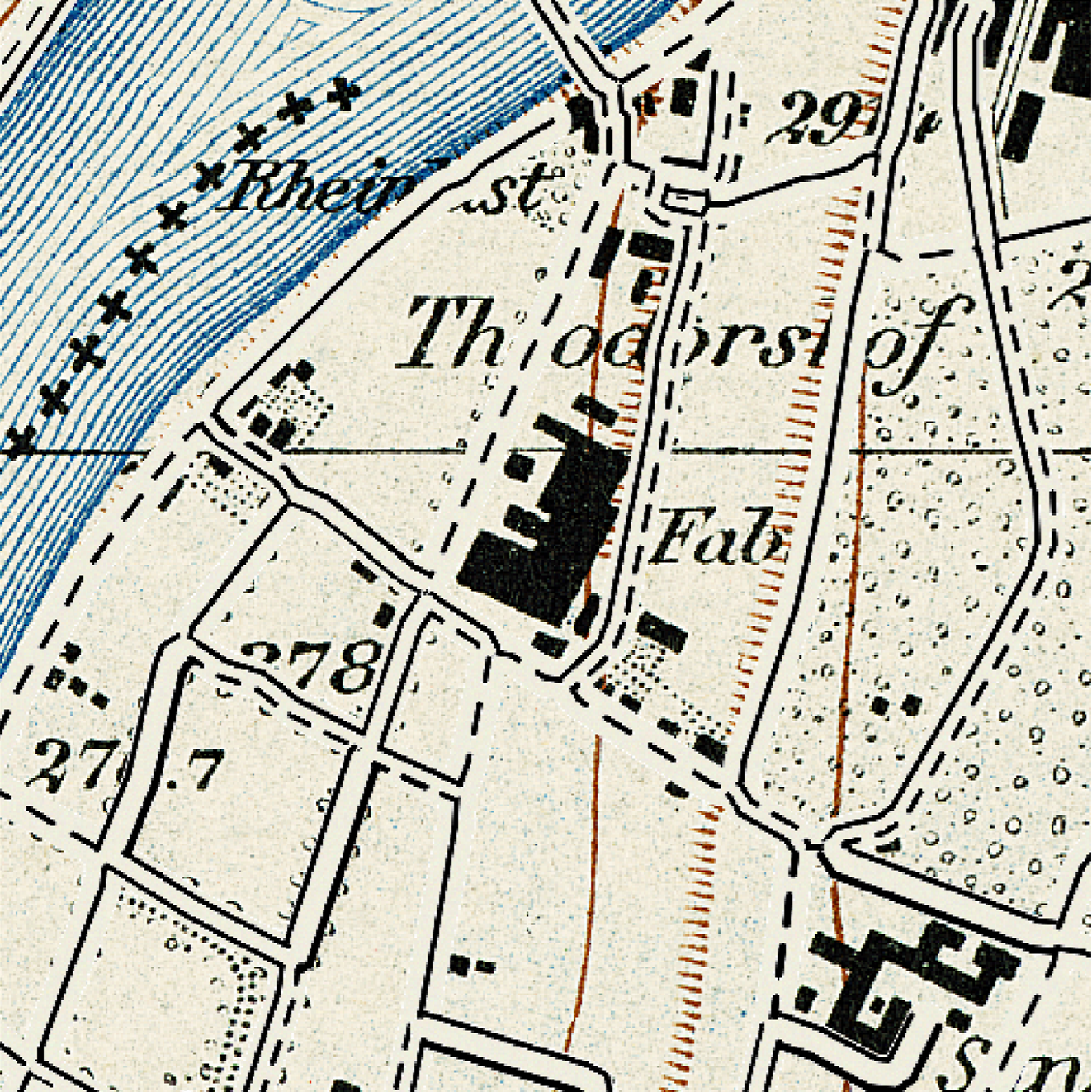}\label{fig:siegfried_synthetic}}}%
    \qquad
    \subfloat[Synthetic labels]{{\includegraphics[width=0.45\linewidth]{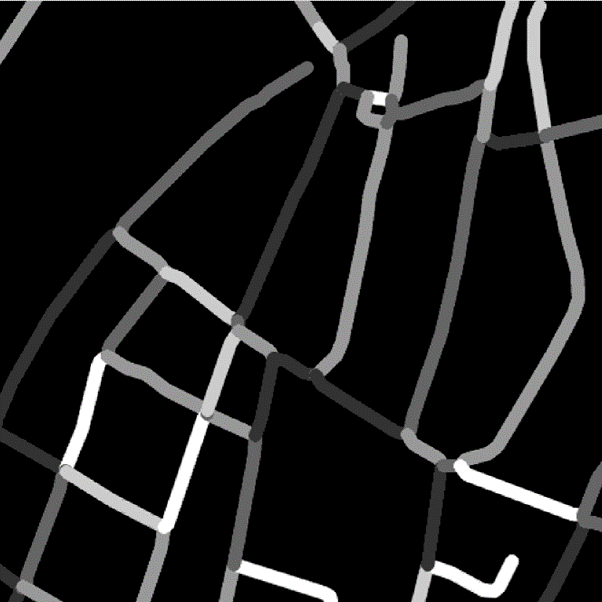} \label{fig:labels_siegfried_synthetic}}}%
    \caption{Creation of synthetic training data. Geodata © Swisstopo\footref{swisstopo}.}
    \label{fig:synthetic_data}
\end{figure}

Given the symbolization of the five road classes and the predicted and vectorized road geometries (Figure~\ref{fig:vectorized_predictions}) for the corresponding \textit{Siegfried Map} sheet (Figure~\ref{fig:original_siegfried}), we aim to train a model that assigns classes to each road. 

First, we randomly assign road classes to each road in the vector dataset. Using these randomly assigned classes, we overlay the original \textit{Siegfried Map} with the vector lines in the corresponding symbolization,  as shown in Figure~\ref{fig:siegfried_synthetic}. This process allows us to create a synthetic \textit{Siegfried Map} with known road class labels. We annotate these labels  (Figure~\ref{fig:labels_siegfried_synthetic}) and then perform the same pre-processing steps described in Section~\ref{sec:seg_preprocessing} for the segmentation. 
With the synthetic \textit{Siegfried Map} and the corresponding synthetic labels, we generate training and validation data suitable for supervised learning.

\subsubsection{Road classification model}
\label{classification_model}
Training a road classification model presents several challenges. Firstly, road segmentation and classification is a difficult task, especially given the limited training data available for our study. To address this, we reuse the pre-trained \textit{Attention ResU-Net} model initially trained on a binary road segmentation task described in Section~\ref{sec:segmentation_model}. We replace its final layer to predict outputs for each road class. Additionally, we employ a hard masking approach to facilitate learning: The classification model is only responsible for predicting road class probabilities, while the road geometries are provided through hard masking. Specifically, we buffer the predicted and post-processed vector geometries with a buffer size of 5 pixels. This mask is then used during training and inference to modify the predicted class likelihoods as follows:
\begin{equation}
\scriptsize
Masking = \begin{Bmatrix}
p(no~road) = 1,~ p(Class ~ i) = 0, ~~ if ~mask= 1\\ 
p(no~road) = 0,~ p(Class ~ i) = p_i, ~~ if ~mask= 0\end{Bmatrix},
\end{equation}
where $p(x)$ refers to the predicted class probability of road class x or the label "no road." Since the binary segmentation model has implicitly learned suitable features for identifying roads of each class, minimal fine-tuning of only two epochs with a constant learning rate of 0.0005 is sufficient.

Another challenge in road classification is ensuring the robustness of the model. Training on synthetic data introduces a distribution shift, as visible in Figure~\ref{fig:synthetic_data}, since synthetic data does not entirely follow the same distribution as the original \textit{Siegfried Map}. Moreover, our framework is based on predicted class probabilities, necessitating that our model is calibrated to produce reliable uncertainty estimations.

To enhance the robustness of our model, we employ several strategies. First, we use the Adam optimizer with a weight decay of 0.00001 for regularization \citep{adam}. Additionally, to improve calibration and add regularization, we apply label smoothing with an epsilon parameter of 0.05 to the cross-entropy loss function \citep{label_smoothing}. We further train an ensemble of models by training 30 models with different initializations of the last layer and varying the order of training images to increase diversity between the ensemble members \citep{Lakshminarayanan2017SimpleEnsembles}. This ensemble approach enhances the model's predictive performance and calibration by addressing epistemic uncertainty. Detailed information regarding training, model selection, and performance can be found in Appendix~\ref{Appendix:classification_model}. As a result of the classification model, we obtain six probabilities for each pixel, five indicating its likelihood of belonging to the corresponding road class and one hard masked probability for not being a road.

\subsubsection{Road class assignment}
\label{sec:statistical_road_class_assignment}
While the output of the classification model allows for pixel-wise road class assignment, further processing is needed to classify the vectorized roads. A straightforward approach would be to assign the road class that is most prominent along a vectorized road segment, where we define a \textit{road segment} as a road between two intersections. 

Although this approach works well for many roads, there may be situations where the road class changes along a \textit{road segment}, leading to incorrect road class assignments. We anoint these points of road class changes along a road segment as \textit{split points}. To accurately identify this \textit {split points}, we developed a methodology based on discretizing and filtering the predicted road class probabilities of the classification model. These predictions implicitly contain already information about potential road class changes due to the fine-grained prediction resolution.

\begin{figure}[h!]
    \centering
    \subfloat[Mean probabilities of the five road classes as a function of distance, including all potential \textit{split points} (shown in gray) and the filtered \textit{split point} (shown in black)]{{\includegraphics[width=1.0\linewidth]{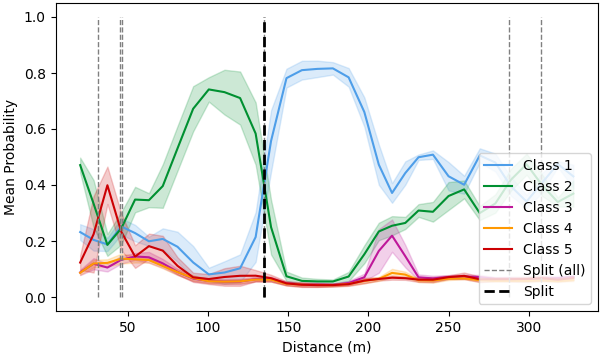}\label{fig:plot_breakpoint}}}\\
    \subfloat[Resulting lines with the split point]{{\includegraphics[width=1.0\linewidth]{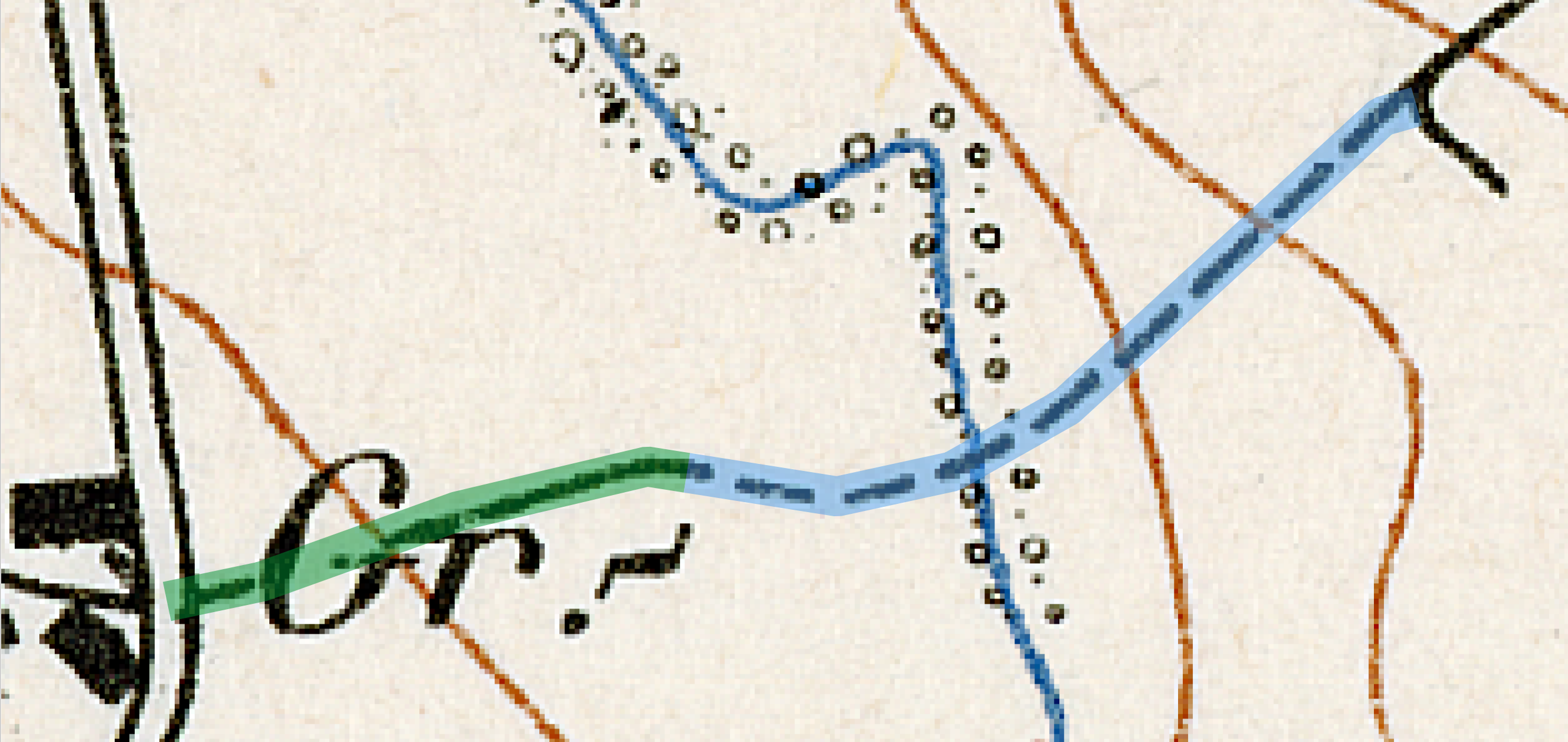} \label{fig:map_breakpoint}}}%
    \caption{Correctly identified \textit{split point} along a \textit{road segment}. Geodata © Swisstopo\footref{swisstopo}.}
\end{figure}

We utilize this information by first dividing the vectorized \textit{road segments} into smaller \textit{road parts}, each with a maximum length of $10~\mathrm{m}$. Each \textit{road part} is then buffered with a $6~\mathrm{m}$ radius, resulting in a set of polygons. Then, the mean value of all pixels within the buffer polygons is calculated for the five road classes. This process results in a discrete probability value for each \textit{road part} of the entire \textit{road segment}. By combining these probabilities with the length of each \textit{road part}, we can plot a line graph that shows the mean probability as a function of distance along a \textit{road segment} for the five road classes, as illustrated in Figure~\ref{fig:plot_breakpoint}.

In this plot, \textit{split points} can be identified at locations where the class with the highest mean probability changes along the \textit{road segment}. This allows us to divide a \textit{road segment} into several sections of different road classes at these potential \textit{split points} (grey, dashed lines). However, the predicted class probabilities can be noisy due to the distribution shifts from the synthetic training data and the original \textit{Siegfried Map}, which may result in unrealistic situations where road classes change frequently, leading to very short sections. Therefore, we introduce an additional filtering mechanism to filter out false positive split points.

More precisely, we define some heuristics for filtering out road sections shorter than $80~\mathrm{m}$ between two \textit{split points}:
\begin{itemize}
\item If the road section before the first \textit{split point} and the road section after the second \textit{split point} have the same assigned road class, all three road sections are merged.
\item If the road section before the first \textit{split point} and the road section after the \textit{split point} point have different assigned road classes, the short road section between the two \textit{split points} is divided in the middle. The first part is merged with the road section before the first \textit{split point}, and the second part is merged with the road section after the second \textit{split point}.
\end{itemize}
We iterate over all road sections and apply this procedure repeatedly until all sections are at least $80~\mathrm{m}$ long. The filtered \textit{split points} (black, dashed lines) are assumed to be true \textit{split points}. Knowing the positions of these filtered \textit{split points}, we can divide the entire \textit{road segment} between two intersections into several \textit{road segment} of different road classes. 

In the final step, all the extracted \textit{road segments} are classified by assigning the road class with the highest mean probability, using zonal statistics with a buffer of $6~\mathrm{m}$. Additionally, we exclude the first and last $20~\mathrm{m}$ of an entire road between two intersections for road class assignment. These segments are within crossroad areas where the predictions are less reliable caused by the distribution shift.

Figure~\ref{fig:map_breakpoint} illustrates an example of a road with a \textit{split point}, while Figure~\ref{fig:plot_breakpoint} displays the corresponding line plot. From a technical standpoint, this process was implemented using the Python libraries Shapely \citep{Shapely} and Rasterstats \citep{rasterstats}. 

\subsection{Evaluation}
In addition to visual assessment, we quantitatively evaluate our approach's resulting road vectorization and classification. We calculate the metrics \textit{Completeness} and \textit{Correctness} to assess the quality of the extracted vector lines. Those metrics are based on the vectorized ground truth (GT) and the classified and vectorized lines \citep{wiedemann2003external}. The computation involves measuring the length of correctly or incorrectly classified lines within a buffer:
\begin{equation}
\scriptsize
\textit{Completeness} = \frac{\text{Length of GT within the buffer of vectorized lines}}{\text{Length of GT}},
\end{equation}

\begin{equation}
\scriptsize
\textit{Correctness} = \frac{\text{Length of vectorized lines within the buffer of GT}}{\text{Length of vectorized lines}}.
\end{equation}

We used a buffer size of five meters for this study. Besides evaluating these metrics for each road class, we calculated a weighted score by weighting the \textit{Completeness} value by the length of each road class in the ground truth. Similarly, we weighted the \textit{Correctness} values by the length of each predicted and vectorized road class, as suggested by \cite{jiao2024novel}.

\section{Experiment, results, and evaluation}
\label{sec:experiment_results_evaluation}
This Section presents and discusses our approach's visual and quantitative results. First, we present the results of extracting and vectorizing roads from historical maps. Next, we visually and quantitatively evaluate and discuss the final results produced by our method. Following this, we discuss the issue of distribution shift between the original \textit{Siegfried Map} and the synthetic road class data used for training and validation. Finally, we analyze the robustness of our approach through a sensitivity analysis of the road assignment algorithm. 

\subsection{Road extraction and vectorization}
\begin{figure}[h]
    \subfloat[\textit{Siegfried Map}]{{\includegraphics[width=0.45\linewidth]{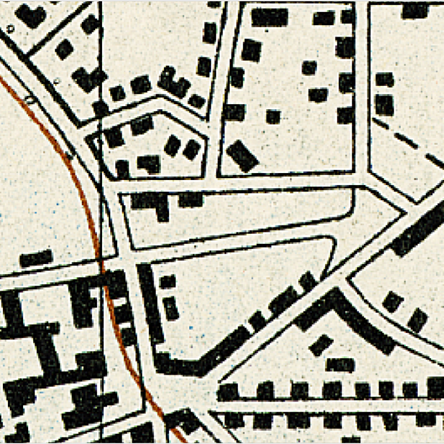}\label{fig:result_seg_1_input}}}%
    \qquad
    \subfloat[Prediction]{{\includegraphics[width=0.45\linewidth]{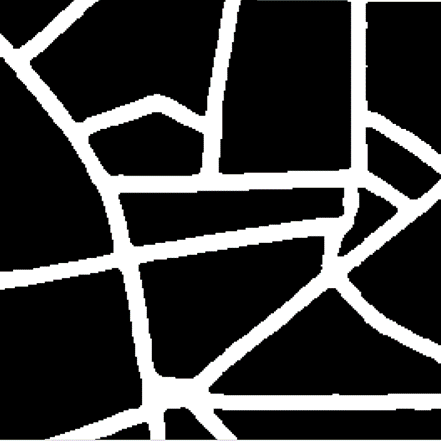} \label{fig:result_seg_1}}}%
    \caption{Prediction of the binary segmentation model \textit{Attention ResU-Net} on the test data. Geodata © Swisstopo\footref{swisstopo}.}
    \label{fig:result_seg_visual}
\end{figure}

A well-performing binary segmentation model for identifying roads is essential for our approach, as the road geometries are derived from the predictions. Figure~\ref{fig:result_seg_visual} presents some visual results of the \textit{Attention ResU-Net} model on the test data. The model accurately segments roads of varying widths and classes and correctly does not frequently identify contour lines or the coordinate grid as roads. Overall, our model demonstrates strong performance in the segmentation task. Quantitative results supporting this assessment can be found in Appendix~\ref{Appendix:generlalization_performance}.

\begin{figure}[h]
    \subfloat[Prediction from the segmentation model]{{\includegraphics[width=0.45\linewidth]{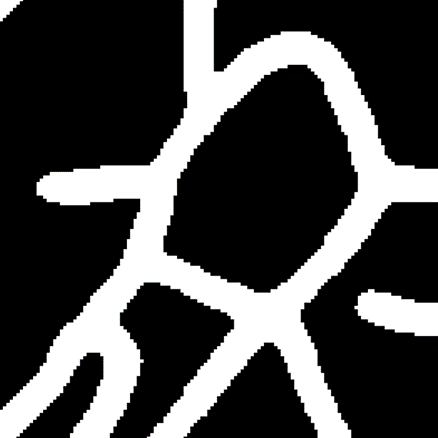}\label{fig:result_vec_2_pred}}}%
    \qquad
    \subfloat[Vectorization of prediction (orange)]{{\includegraphics[width=0.45\linewidth]{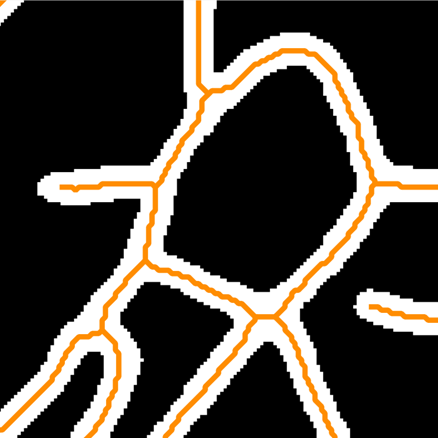} \label{fig:result_vec_2_vec}}}%
    \qquad
    \subfloat[Generalization (red)]{{\includegraphics[width=0.45\linewidth]{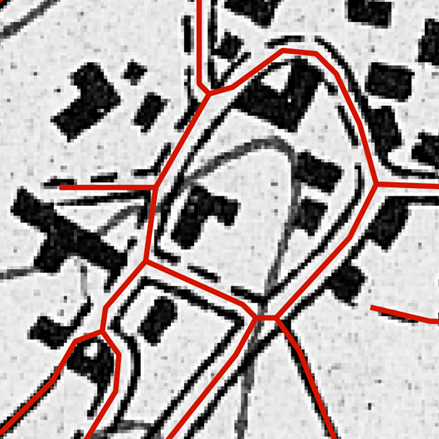}\label{fig:result_vec_2_dp}}}%
    \qquad
    \subfloat[Ground truth (green)]{{\includegraphics[width=0.45\linewidth]{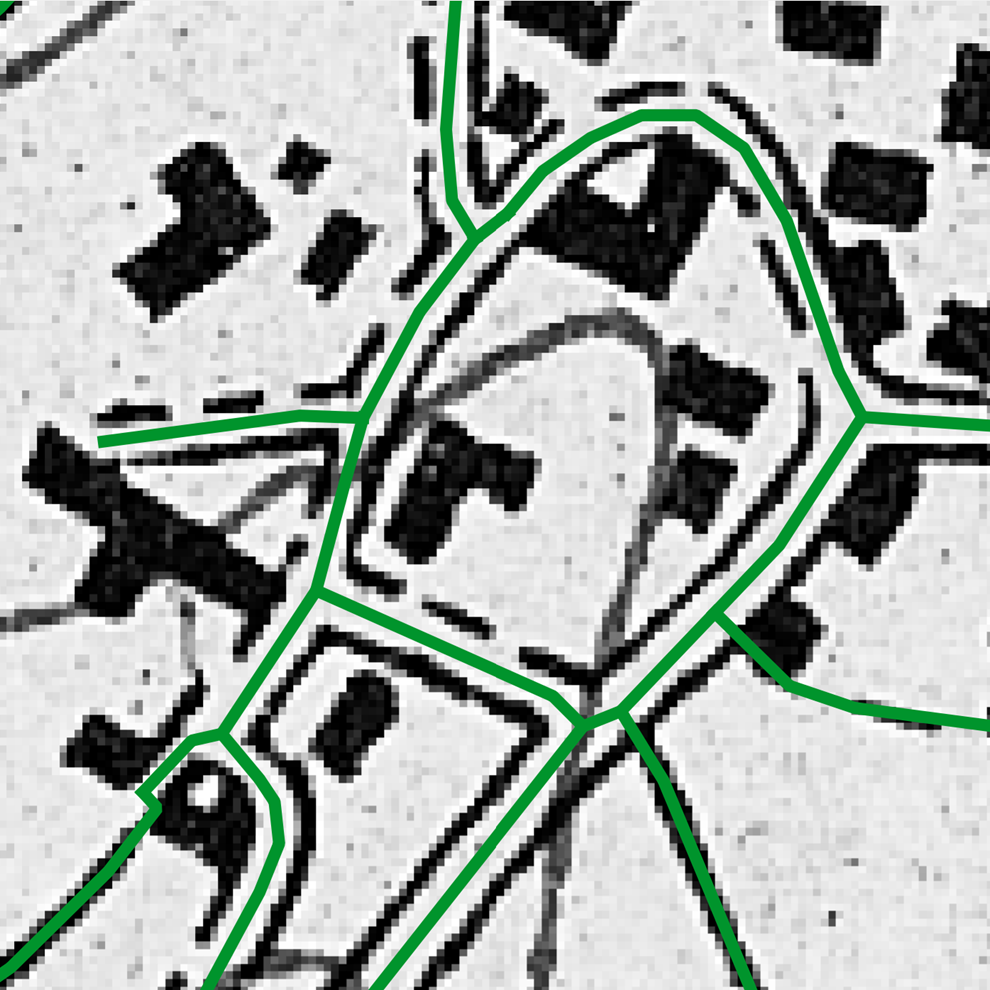} \label{fig:result_vec_2_gt}}}%
    \caption{Visual assessment of the vectroization result with a challenging situation. Geodata © Swisstopo\footref{swisstopo}.}
    \label{fig:result_vec_visual}
\end{figure}

Figure~\ref{fig:result_vec_visual} illustrates the following vectorization process. First, the binary segmentation model predictions undergo morphological operations, and skeletonization followed by vectorization, as shown in Figure~\ref{fig:result_vec_2_vec}. The resulting vector dataset approximates the axes of the extracted roads. However, the vector dataset contains zigzag lines due to noise in the predictions and excessive support points per road. We applied the Douglas-Peucker algorithm to address this \citep{douglas_peucker}, resulting in a smoother vector dataset, as presented in Figure~\ref{fig:result_vec_2_dp}.
Comparing the resulting vector dataset with the ground truth in Figure~\ref{fig:result_vec_2_gt}, we see that our approach successfully vectorizes roads. However, limitations remain. For instance, skeletonization can lead to inaccuracies at road intersections due to changes in the center of mass where three or more road segments meet, or, the generalization can lead to an inaccurate road axis, such as the driveway in the middle left. Additionally, some inaccuracies are introduced by the segmentation model in challenging situations, such as the dashed road symbol at the bottom right near the house, where the model incorrectly inferred that the road does not continue.

\begin{figure}[h]
    \subfloat[Generalized road geometries with filtered misidentified roads (red) and not filtered roads (green)]{{\includegraphics[width=0.45\linewidth]{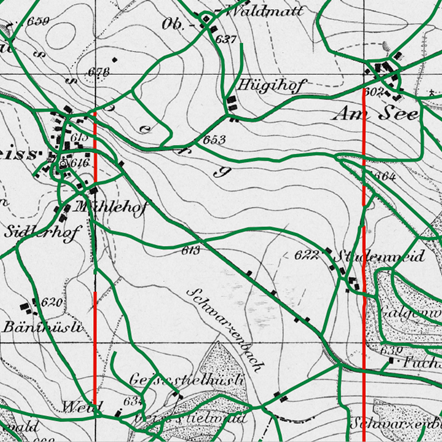}\label{fig:result_3_coord_grid_filter}}}%
    \qquad
    \subfloat[Ground truth (blue)]{{\includegraphics[width=0.45\linewidth]{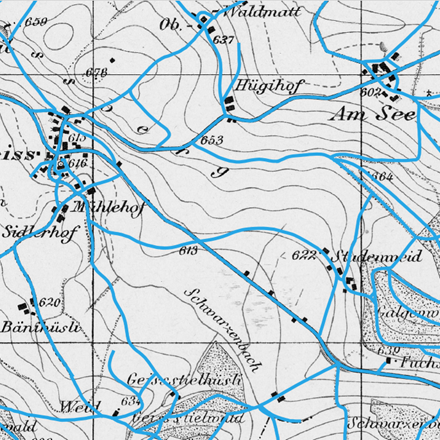} \label{fig:result_3_coord_grid_filter_gt}}}%
    \caption{Coordinate grid filtering in challenging situations. Geodata © Swisstopo\footref{swisstopo}.}
    \label{fig:coord_grid_filter}
\end{figure}

Despite pre-training our binary segmentation model on \textit{Swiss Map}, there are still instances where the model incorrectly classifies the coordinate grid as a road. This issue arises because the coordinate grid uses symbology similar to road class 2 (Figure~\ref{fig:road_classes}). Our analytical coordinate grid filter significantly enhances the quality of the vectorized dataset by accurately removing these incorrectly extracted roads, as demonstrated in Figure~\ref{fig:coord_grid_filter}.

\begin{table}[h!]
\centering
\begin{tabular}{lcc}
\toprule
\textbf{Class} & \textbf{\textit{Completeness} [\%]} & \textbf{\textit{Correctness} [\%]}\\
\midrule
Class 1 & 86.06 & 94.01  \\
Class 2 & 94.23 & 92.58  \\
Class 3 & 94.76 & 88.00  \\
Class 4 & 79.36 & 93.72  \\
Class 5 & 89.51 & 72.89 \\
\midrule
Weighted & 91.01 & 91.64\\
\bottomrule
\end{tabular}
\caption{Final scores with a segmentation interval $\delta = 10~\mathrm{m}$, a minimal line length $l=80~\mathrm{m}$, and buffer size $\beta  = 6~\mathrm{m}$ for the zonal statistics.}
\label{tab:class_scores}
\end{table}

\subsection{Quantitative and visual assessment}
The evaluation of our approach resulted in a weighted score of 91.01\% for \textit{Completeness} and 91.64\% for \textit{Correctness}. For Class 2, the most frequent class in the \textit{Siegfried Map}, we achieved scores of over 94\% and 92\% using only the pure road geometries and symbolization as training data. Class 5 was the most challenging to classify correctly, probably because the line width of the symbol varies from map sheet to map sheet, due to the inherent quality issues of historical maps. Our implementation based on OpenCV includes a random element for the line widths, the spacing between double line symbols, and the dashing, in order to produce synthetic data with a certain level of variability. Since the \textit{Siegfried Map} is a hand-made map, all symbolizations are subject to a certain variability, and thus, this random component allows us to further reduce the distribution shift. Detailed evaluations for all road classes are shown in Table~\ref{tab:class_scores}.

\begin{figure}[h]
    \subfloat[Prediction]{{\includegraphics[width=0.45\linewidth]{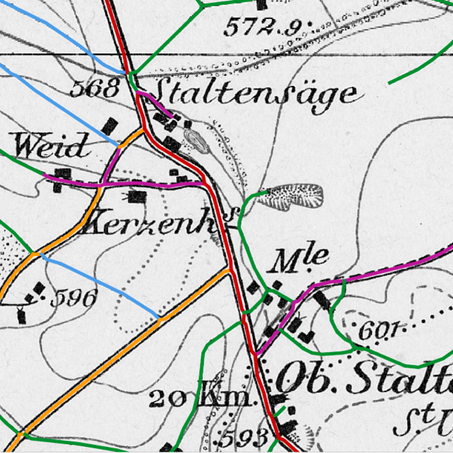}\label{fig:result_1_pred}}}%
    \qquad
    \subfloat[Ground truth]{{\includegraphics[width=0.45\linewidth]{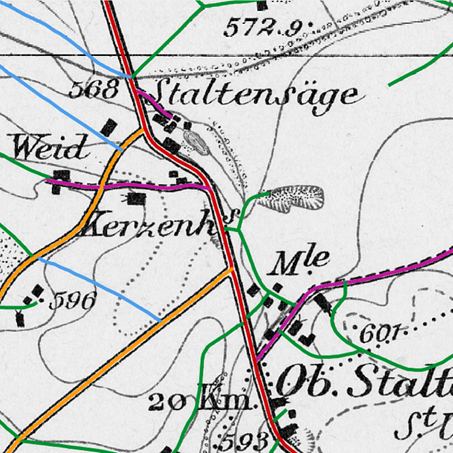} \label{fig:result_1_gt}}}%
    \qquad
    \subfloat[Prediction]{{\includegraphics[width=0.45\linewidth]{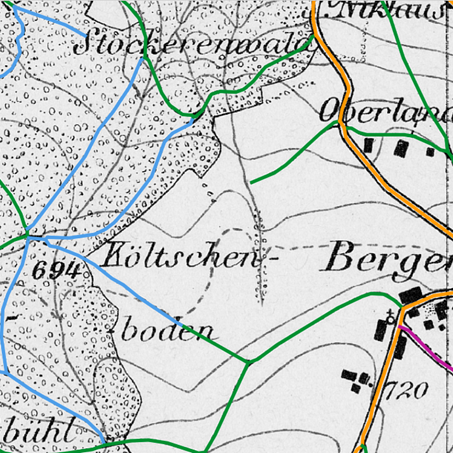}\label{fig:result_2_pred}}}%
    \qquad
    \subfloat[Ground truth]{{\includegraphics[width=0.45\linewidth]{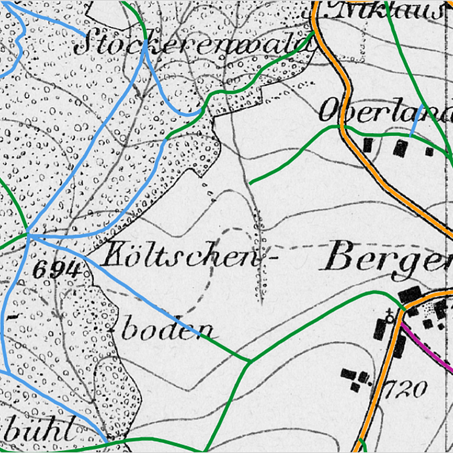} \label{fig:result_2_gt}}}%
    \qquad
    \subfloat[Prediction]{{\includegraphics[width=0.45\linewidth]{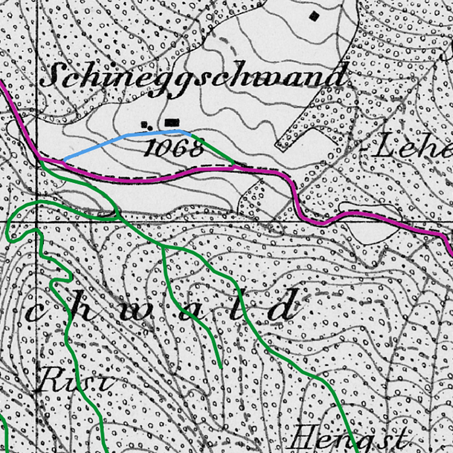}\label{fig:result_5_pred}}}%
    \qquad
    \subfloat[Ground truth]{{\includegraphics[width=0.45\linewidth]{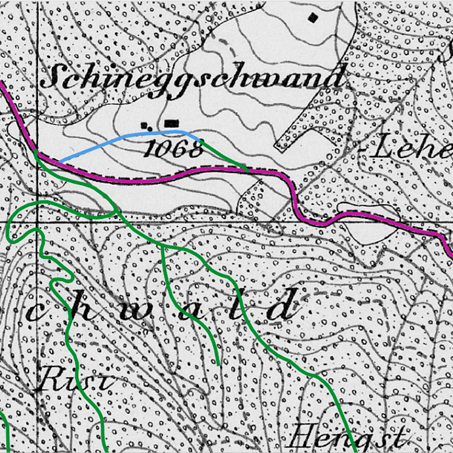} \label{fig:result_5_gt}}}%
    \caption{Visual assessment of the final result. The \textit{Siegfried Map} is shown in grayscale and the vectorized labeled roads in colour, whereby class 1 is coloured blue, class 2 green, class 3 purple, class 4 orange and class 5 red. Geodata © Swisstopo\footref{swisstopo}.}
    \label{fig:result_visual_good}
\end{figure}

\begin{figure}[h]
    \subfloat[Prediction]{{\includegraphics[width=0.45\linewidth]{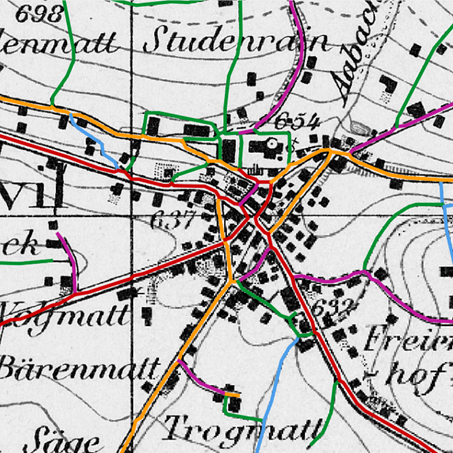}\label{fig:result_3_pred}}}%
    \qquad
    \subfloat[Ground truth]{{\includegraphics[width=0.45\linewidth]{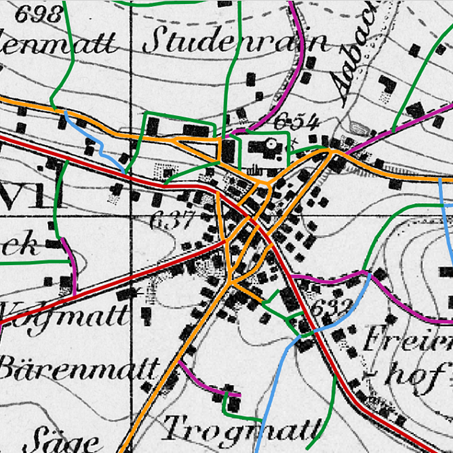} \label{fig:result_3_gt}}}%
    \qquad
    \subfloat[Prediction]{{\includegraphics[width=0.45\linewidth]{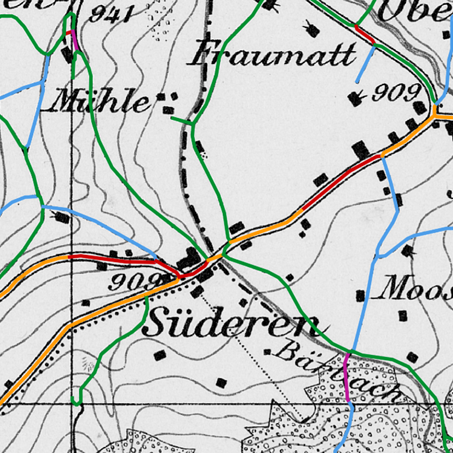}\label{fig:result_4_pred}}}%
    \qquad
    \subfloat[Ground truth]{{\includegraphics[width=0.45\linewidth]{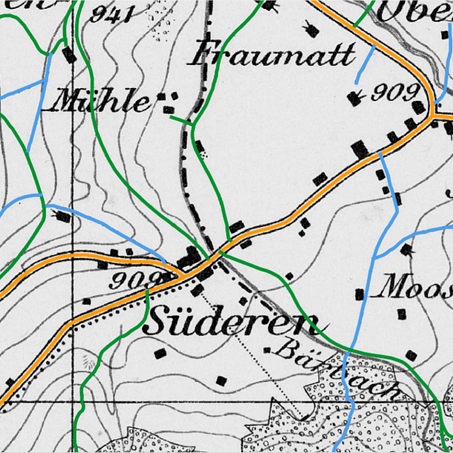} \label{fig:result_4_gt}}}%
    \caption{Visual assessment of the final result with challenging situations. The \textit{Siegfried Map} is shown in grayscale and the vectorized labeled roads in colour, whereby class 1 is coloured blue, class 2 green, class 3 purple, class 4 orange and class 5 red. Geodata © Swisstopo\footref{swisstopo}.}
    \label{fig:result_visuel_bad}
\end{figure}

Figure~\ref{fig:result_visual_good} shows the results where our approach worked well, while challenging situations are shown in Figure~\ref{fig:result_visuel_bad}. The most accurate results were achieved outside settlements, while the performance in dense areas is lower, as shown in Figure~\ref{fig:result_3_pred}. Taking a closer look at the painting process, we can observe that we first need to overdraw all roads with a width of $13~$pixels. This is necessary because the roads in class 5 are this wide, and we need to cover them completely. Since narrower roads are often present in villages, and buildings were often built right up to the roads at that time, we also end up overdrawing many buildings and can no longer create synthetic roads that reach up to the buildings. This results in a particularly large distribution shift in this situation, which likely leads to lower performance. Furthermore, roads within a settlement are often shorter than $80~\mathrm{m}$ and have consequently fewer pixels available for classification than longer roads. Therefore, outliers in the model output have a greater influence on the classification. Another reason for the inferior performance on very short roads (less than $40~\mathrm{m}$ in length) is that we are unable to crop the intersection areas for classification. These areas pose two main challenges: First, the quality of the painted intersections is often inferior, resulting in a significant distribution shift. Second, the vectorization algorithm assigns the intersection point to the center of mass of the intersection rather than the geometrically correct intersection point. Consequently, we end up training on intersection areas with inaccurate synthetic geometries.

Figure~\ref{fig:result_4_pred} illustrates an example of difficulties classifying classes 4 and 5 overland. These symbols are very similar and exhibit slight variations from map to map, which causes the classifier to struggle to distinguish between the two classes. A similar issue is observed with classes 1, 2, and 3. In Figure~\ref{fig:result_4_pred}, the symbolizations for classes 1 and 2 consist of individual lines overlaid on a forest texture. However, in the synthetic training data, roads in forests have a broad yellowish background for covering completely the original roads on the \textit{Siegfried Map}. This results in a particularly large distribution shift in forest areas, making it unsurprising that the classifier occasionally confuses these classes with class 3.

\subsection{Synthetic training data: Effect of distribution shift}
\label{sec:effect_dist_shift}
In this Section, we analyze the impact of distribution shift on the performance of our road classification model. Although using synthetic labeled data reduces the cost of generating training data, the distribution shift between synthetic and real data often challenges the model's ability to generalize well \citep{Zhang2021DeepSL}. Figure~\ref{fig:synthetic_data} illustrates the limitations in the quality of synthetic data, highlighting the risk that original \textit{Siegfried Map} inputs may be problematic for the network to generalize due to being out-of-distribution. Additionally, reliable uncertainty estimations in such settings are challenging \citep{Zhang2021DeepSL}.

\begin{figure}[h]
\centering
    \subfloat[$F1~Score$ ($\uparrow$)]{{\includegraphics[width=0.45\textwidth]{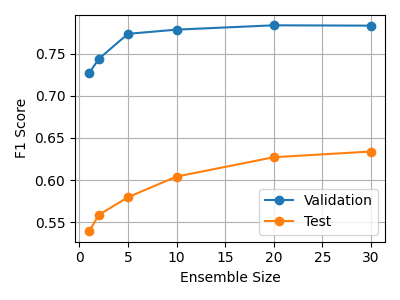}\label{fig:ensemble_size_F1}}}%
    \qquad
    \subfloat[$Brier~Score$ ($\downarrow$)]{{\includegraphics[width=0.45\textwidth]{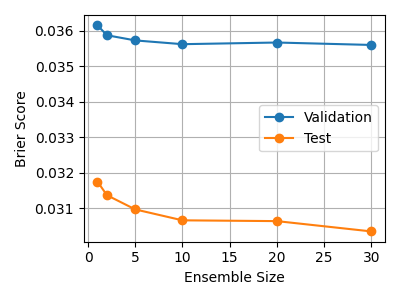} \label{fig:ensemble_size_brier}}}%
    \caption{Evaluation metrics dependent on the ensemble size for validation and test data. Validation data is synthtetic, while test data refers to the original \textit{Siegfried Map}.}
    \label{fig:ensemble_size}
\end{figure}

We investigated the effect of ensembling on improving the robustness and calibration of the model. Figure~\ref{fig:ensemble_size} shows the $F1~Score$ and $Brier~Score$ for both the validation and test sets. Increasing the ensemble size enhances predictive performance and improves the quality of the predicted class probabilities. The effect is more pronounced on the test data than on the synthetic validation data due to the higher epistemic uncertainty in the test data caused by the distribution shift. Since ensemble members may converge to different modes of the loss function, each member can be seen as a Monte Carlo sample from the posterior distribution, where averaging predictions of different members results in more reliable predictions for regions in the input space lacking training data \citep{baysian_deep_learning_modes, baysian_modes}. 

Surprisingly, the $Brier~Score$ is lower for the synthetic validation data. This pattern should be interpreted with care since the $Brier~Score$ metric includes also the majority class of "no road" pixels. This class is predicted by the hard masking mechanism, meaning that it is not affected by the distribution shift. More results and the definition of the evaluation metrics can be found in Appendix~\ref{Appendix:classification_model_ensemble_size} and \ref{Appendix:evaluation_metrics}.

\subsection{Sensitivity analysis for hyperparameters of the road class assignment}
\label{Sec:results_sensitivity_analysis}
The split point detection with the consecutive road class assignment has several hyperparameters that must be selected before applying the framework. These hyperparameters are the discretization or \textit{segmentation interval} $\delta$ to map the predicted probabilities to the road lines; the \textit{minimum line length} $l$ that a section must have between two split points; and the \textit{buffer size} $\beta$ used in the zonal statistics. To evaluate the robustness of our road class assignment, we conducted a sensitivity analysis. This was achieved by varying specific parameters and assessing the impact on classification performance. The results are presented in Table~\ref{tab:sensitivity_analysis}.

For the \textit{segmentation interval} $\delta$, there exists a trade-off between slightly higher scores and the processing time. We therefore suggest an interval of $10~\mathrm{m}$, although the scores are marginally higher at $5~\mathrm{m}$. For a large-scale application, this post-processing step could also be further optimized.

The minimum line length parameter, denoted as $l$, appears to have an optimal range between $40~\mathrm{m}$ and $120~\mathrm{m}$. Our initial evaluation defined this parameter as $80~\mathrm{m}$. This parameter influences the method's sensitivity to a certain extent; thus, it can partially suppress noise. However, it must be noted that it may also suppress true split points.

While our classification model predicts all roads as 13~pixel lines, equivalent to $16.25~\mathrm{m}$ in width, we discovered that predictions near the road edges tend to degrade classification performance. Consequently, we introduced an additional hyperparameter, the buffer size $\beta$. This parameter appears to be optimally set around $6~\mathrm{m}$. The primary issue likely lies in the fact that for road classes 3, 4, and 5, classes 1 or 2 are more likely to be predicted at the edges of roads, with the correct classification of 3, 4, or 5 near the road axis. This can be explained by the similarity in symbology. Conversely, for classes 1 and 2, factors such as texture, buildings, or fonts may exert a greater influence towards the edge of the prediction area. A larger buffer size covering more predicted pixels contributes to a larger sample size for determining the mean value. This situation presents two contrasting arguments - one advocating for a wider buffer size and the other for a narrower one. The equilibrium between these opposing factors appears to be reached with a buffer size of approximately $6~\mathrm{m}$.

Surprisingly, the results are very similar despite the different hyperparameters; in fact, all weighted values are about 90\%. Overall, these results demonstrate that our approach is very robust regarding parameter choices, making it feasible to rely purely on synthetic labeled data, and potentially generalizable to other historical map or even remote sensing datasets.

\begin{table}[h!]
\scriptsize
\centering
\begin{tabular}{lllccc}
\toprule
$\boldsymbol{\delta}$ & $\boldsymbol{l}$ & $\boldsymbol{\beta}$ & & \textbf{\textit{Comp.} [\%]} & \textbf{\textit{Corr.} [\%]}\\
\midrule
\textbf{5m}  & \textit{10m} & \textit{6m} & Class 1  & 86.08 & 94.29\\
	& & & Class 2  & 94.30 & 92.66\\
	& & & Class 3  & 95.31 & 88.06\\
	& & & Class 4  & 80.19 & 94.59\\
	& & & Class 5  & 89.51 & 73.41\\
	& & & Weighted & 91.20 & 91.83\\
    & & & & & \\
\textbf{20m} & \textit{10m} & \textit{6m} & Class 1  & 85.78 & 93.80\\
	& & & Class 2  & 94.27 & 92.44\\
	& & & Class 3  & 94.39 & 88.33\\
	& & & Class 4  & 79.31 & 93.45\\
	& & & Class 5  & 89.53 & 72.43\\
	& & & Weighted & 90.92 & 91.53\\
    & & & & & \\
\midrule
\textit{10m} & \textbf{40m} & \textit{6m} & Class 1  & 83.15 & 93.45\\
&	& & Class 2  & 93.68 & 91.57\\
&	& & Class 3  & 93.61 & 84.46\\
&	& & Class 4  & 77.79 & 91.53\\
&	& & Class 5  & 89.56 & 68.95\\
&   & & Weighted & 89.77 & 90.09\\
&   & & & & \\
\textit{10m} & \textbf{120m}& \textit{6m} & Class 1  & 85.57 & 92.86\\
&	& & Class 2  & 93.83 & 92.48\\
&	& & Class 3  & 94.69 & 87.92\\
&	& & Class 4  & 79.67 & 94.32\\
&	& & Class 5  & 91.26 & 74.35\\
&	& & Weighted & 90.75 & 91.43\\
&   & & & & \\
\midrule
\textit{10m} & \textit{80m} & \textbf{4m}  & Class 1  & 84.85 & 94.03 \\
& &     & Class 2  & 94.14 & 92.61 \\
& &     & Class 3  & 94.39 & 86.92 \\
& &     & Class 4  & 79.54 & 93.00 \\
& &     & Class 5  & 90.37 & 69.74 \\
& &     & Weighted & 90.68 & 91.29 \\
& &     & & & \\
\textit{10m} & \textit{80m} & \textbf{10m} & Class 1  & 87.17 & 91.78\\
& &     & Class 2  & 93.32 & 91.20\\
& &     & Class 3  & 94.85 & 88.12\\
& &     & Class 4  & 73.73 & 95.06\\
& &     & Class 5  & 84.57 & 78.81\\
& &     & Weighted & 90.16 & 90.79\\
& &     & & & \\
\bottomrule
\end{tabular}
\caption{Results of the sensitivity analysis for the hyperparameter of the
split point detection: segmentation interval $\delta$, minimum line length $l$, and buffer size $\beta$. The final results are shown in Table~\ref{tab:class_scores} and were calculated with $\delta = 10\text{m}$, $l = 80\text{m}$, and $\beta = 6\text{m}$}
\label{tab:sensitivity_analysis}
\end{table}

\section{Discussion}
\label{sec:discussion}
The results in Section~\ref{sec:experiment_results_evaluation} demonstrate that our proposed approach accurately vectorizes and classifies roads in historical maps. We utilize deep learning to extract roads from historical maps (Figure~\ref{fig:result_seg_visual}), which are then post-processed and vectorized (Figure
~\ref{fig:result_vec_visual}). The resulting vector dataset is used to generate synthetic road class labeled data by leveraging symbol painting (Figure~\ref{fig:synthetic_data}). We then employ probabilistic deep learning by training on the synthetic labeled dataset and predicting accurate road class probabilities. Subsequently, we analyze the predicted probabilities along the \textit{road segments} to identify \textit{split points}, where the road class changes, allowing accurate road class assignment. Our method has proven to be robust and high-performing based on both visual assessments (Figure~\ref{fig:result_visual_good}) and quantitative assessments (Tables~\ref{tab:class_scores} and \ref{tab:sensitivity_analysis}), despite the problematic distribution shift between the synthetic data and the original \textit{Siegfried Map}.

This work presents innovations in vectorization and classification of roads and other symbols in historical maps. Our approach yields a vector dataset with class labels, distinguishing it from previous research \citep{chiang2009automatic, jiao2022fast}. The most closely related work is by \cite{jiao2024novel}, which made significant contributions by utilizing symbol painting for road classification without the need for expensive, class-labeled training data. This method involves classifying roads using the painting function to symbolize each extracted road segment directly. Then, the most appropriate road class is chosen by finding the symbology that minimizes the difference between the symbolized road and the input \textit{Siegfried Map}. We believe that our approach extends the previous work by using symbol painting to generate synthetic data for conducting probabilistic road classification based on deep learning. Our sophisticated road class assignment further allows us to accurately find changes in road class along a route. Our approach particularly excels in producing superior results for classes where symbol painting-based classification is challenging, such as road class 1 (Figure~\ref{fig:road_classes}), where the placement of each dash is problematic. We address this issue through the translation equivariance of neural networks, whereby the exact position of a dash in a road does not influence the classification, compared to template matching-based approaches.

Although we used the \textit{Siegfried Map} for our study, our method should be applicable to other historical map series, as road symbology is often similar. Slight modifications to the symbol painting should be sufficient for applying the method. The chosen approach could likely also be applied to other line elements, such as tram or railway lines, streams or rivers. Future research could also consider utilizing a modified version of our framework for remote sensing applications, such as the extraction and classification of line features from satellite images.

Despite our approach performing well, especially given the small amount of unlabeled training data, some limitations can still be improved. We suspect that the distribution shift between synthetic training and test data is our approach's primary source of error: Since the widest roads have a width of 13 pixels, we need to paint over all roads in the synthetic training data with that width to cover them completely. This results in a change in the distribution between artificial and real data in more densely populated areas with narrow roads and buildings close to the roads. In these areas, roads classified as 1 or 2 will never have buildings directly aligned with the road. Therefore, our model still has problems in these regions, as visible in Figure~\ref{fig:result_3_pred}. 
Secondly, the vectorization process still has a lot of potential. So far, only a pixel-by-pixel vectorization of the skeletonization with subsequent generalization has been implemented. A more sophisticated method, such as the algorithm of \cite{vec_satellite_seg} for the topologically correct vectorization of roads from binary segmented satellite images or the framework developed by \cite{vec_line_drawings} for the vectorization of hand-drawn plans could further improve our approach.

Additionally, our approach can easily make use of other neural network architectures. Interesting is the application of transformer models for road extraction \citep{Vaswani2017, Dosovitskiy2020AnScale, Zhang2024}. Despite the smaller inductive bias of these architectures compared to convolutional neural networks, transformers require large amounts of data to achieve superior performance \citep{Dosovitskiy2020AnScale}. Recent studies have explored parameter-efficient fine-tuning of transformer models using Low-Rank Adaptation (LoRA) \citep{Hu2021LoRA}, also used for enhancing model calibration and performance \citep{LoRA_ensemble}. By pretraining a segmentation transformer on various cartographic or non-cartographic tasks using self-supervised learning \citep{SelfSupervised}, and then fine-tuning it with historical maps, one can potentially achieve superior results.

\section{Conclusion}
Our study introduces an innovative approach for classifying and converting roads from historical maps into vector format. The method employs \textit{cascaded training} on a neural network for road segmentation, followed by post-processing and vectorization. We create synthetic road class-labeled training data to address the challenge of expensive manual labeling. Our approach is applicable to classifying other symbols on historical maps.

We showcase the efficiency and performance of our framework using the Swiss \textit{Siegfried Map}. Through visual assessments, quantitative evaluations, and sensitivity analysis, we conclude that our method enhances accuracy and robustness compared to existing approaches. 

Our key contributions are as follows: firstly, we demonstrate the use of symbol painting to generate synthetic labeled training data, achieving sufficient quality to train neural networks for road classification. 
Secondly, we developed a sophisticated classification algorithm that takes the predicted road class probabilities as input and accurately classifies road segments, even when the road class changes along a segment.
Finally, our comprehensive framework, including advanced data processing such as coordinate grid filtering, yields promising results, as evidenced by visual and quantitative assessments. This leads to cost and time savings through the automation of manual work. The vectorized and classified roads can be utilized for various studies,  including the analysis of road network evolution, emerging economies,  urban development, and the design of sustainable transport infrastructures.

\section*{CRediT author statement}
\textbf{Dominik J. Mühlematter}: Conceptualization, Data curation, Methodology, Formal analysis, Software, Validation, Visualization, Writing -- original draft, Writing -- review \& editing.
\textbf{Sebastian Schweizer}: Conceptualization, Data curation, Methodology, Formal analysis, Software, Validation, Visualization, Writing -- original draft, Writing -- review \& editing.
\textbf{Chenjing Jiao}: Conceptualization, Methodology, Resources, Project administration, Supervision, Writing -- review \& editing.
\textbf{Xue Xia}: Conceptualization, Methodology, Writing -- review \& editing.
\textbf{Magnus Heitzler}: Conceptualization, Writing -- review \& editing.
\textbf{Lorenz Hurni}: Conceptualization, Resources, Funding acquisition, Supervision, Writing -- review \& editing.

\section*{Declaration of competing interest}
The authors declare that they have no known competing financial interests or personal relationships that could have appeared to influence the work reported in this paper.

\section*{Declaration of generative AI and AI-assisted technologies in the writing process}
During the preparation of this work the authors used the service Grammarly in order to improve the readability of the manuscript. After using this service, the authors reviewed and edited the content as needed and take full responsibility for the content of the published article.

\bibliographystyle{cas-model2-names}

\bibliography{cas-refs}


\newpage

\appendix
\section*{Appendix}
\section{Data}
\label{Appendix:data}
This Section provides further details about the datasets we used for this study.

\subsection{Siegfried Map}
\label{Appendix:data_siegfried}
The \textit{Siegfried Map} sheets are available as GeoTIFFs with a resolution of 1.25m per pixel in the reference frame CH1903 (EPSG: 21781). Table~\ref{tab:siegfried_data} presents the region and year of each of the map sheets used for training, validation, or testing. Note that geometries are only available for the city of Zurich for the training data; therefore, only parts of the four training sheets were used for training. In order to mask the \textit{Siegfried Map} sheets without complete ground truth coverage, we downloaded the city boundary from the geoportal of the city of Zurich \footnote{\url{https://www.stadt-zuerich.ch/geodaten/download/95}}.

\begin{table}[h!]
\centering
\begin{tabular}{lccc}
\toprule
\textbf{Split} & \textbf{Sheet Number} & \textbf{Region} & \textbf{Year}\\
\midrule
Train & 158 & Schlieren & 1940  \\
Train & 159 & Schwamendingen & 1940  \\
Train& 160 & Birmensdorf & 1940  \\
Train & 161 & Zürich & 1940  \\
Validation & 017 & Rheinfelden & 1940 \\
Test & 199 & Ruswil & 1941 \\
Test & 385 & Schangnau & 1941 \\
\bottomrule
\end{tabular}
\caption{Split, region, and year of each map sheet of the \textit{Siegfried Map}.}
\label{tab:siegfried_data}
\end{table}

\subsection{Swiss Map}
\label{Appendix:data_swissmap}
\begin{figure}[h!]
    \includegraphics[width=1.0\linewidth]{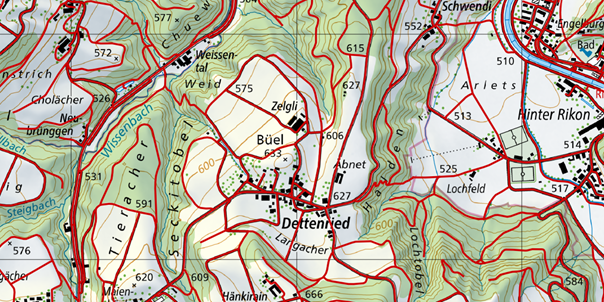}
    \caption{Example excerpt from the \textit{Swiss Map Raster 25} with overlaid roads from the \textit{Swiss Map Vector 25} (red). Geodata © Swisstopo\footref{swisstopo}.}
    \label{fig:swissmap_raster_vector}
\end{figure}

We utilized the Swiss national map to pretrain our binary segmentation model. Specifically, we used the raster dataset \textit{Swiss Map Raster 25} with a resolution of 1.25m per pixel as input data, along with the road geometries from the \textit{Swiss Map Vector 25} dataset. Both datasets are georeferenced in the CH1903+ reference frame (EPSG: 2056). An overview of the map sheets can be found in Table~\ref{tab:swissmap_data}; Figure~\ref{fig:swissmap_raster_vector} shows an example excerpt from the \textit{Swiss Map Raster 25} with overlaid roads from the \textit{Swiss Map Vector 25} in red.

\begin{table}[h!]
\centering
\begin{tabular}{lccc}
\toprule
\textbf{Split} & \textbf{Sheet Number} & \textbf{Region} & \textbf{Year}\\
\midrule
Train & 1052 & Andelfingen& 2019  \\
Train & 1053 & Frauenfeld& 2019\\
Train& 1072& Winterthur& 2019\\
Train & 1073& Wil& 2019\\
Train & 1125 & Chasseral& 2021\\
Train & 1130& Hochdorf& 2021\\
Train & 1131& Zug& 2021\\
Train & 1144 & Val de Ruz & 2020\\
Train & 1145& Bieler See& 2021\\
Train & 1150& Luzern& 2021\\
Train & 1151& Rigi& 2021\\
Train & 1164& Neuchâtel& 2020\\
Train & 1165& Murten / Morat& 2020\\
Validation & 1166& Bern & 2021\\
Validation & 1167& Worb & 2021\\
Train & 1184& Payerne & 2020\\
Train & 1185& Fribourg / Freiburg & 2020\\
Validation & 1186 & Schwarzenburg & 2021\\
Validation & 1187 & Münsingen & 2021\\
\bottomrule
\end{tabular}
\caption{Split, region, and year of each map sheet of the \textit{Swiss Map}.}
\label{tab:swissmap_data}
\end{table}

\section{Binary segmentation}
\label{Appendix:binary_segmentation}
This section provides further details about the binary segmentation model used for extracting roads from historical maps. It covers model selection and evaluation, hyperparameter tuning, training, and investigations into data augmentation. The metrics used in this chapter to assess the models' performance are specified in detail in Appendix \ref{Appendix:evaluation_metrics}.

\subsection{Baseline model architecture}
\label{Appendix:baseline_architecture}
\begin{figure*}[h]
    \centering
    \includegraphics[width=1.0\textwidth]{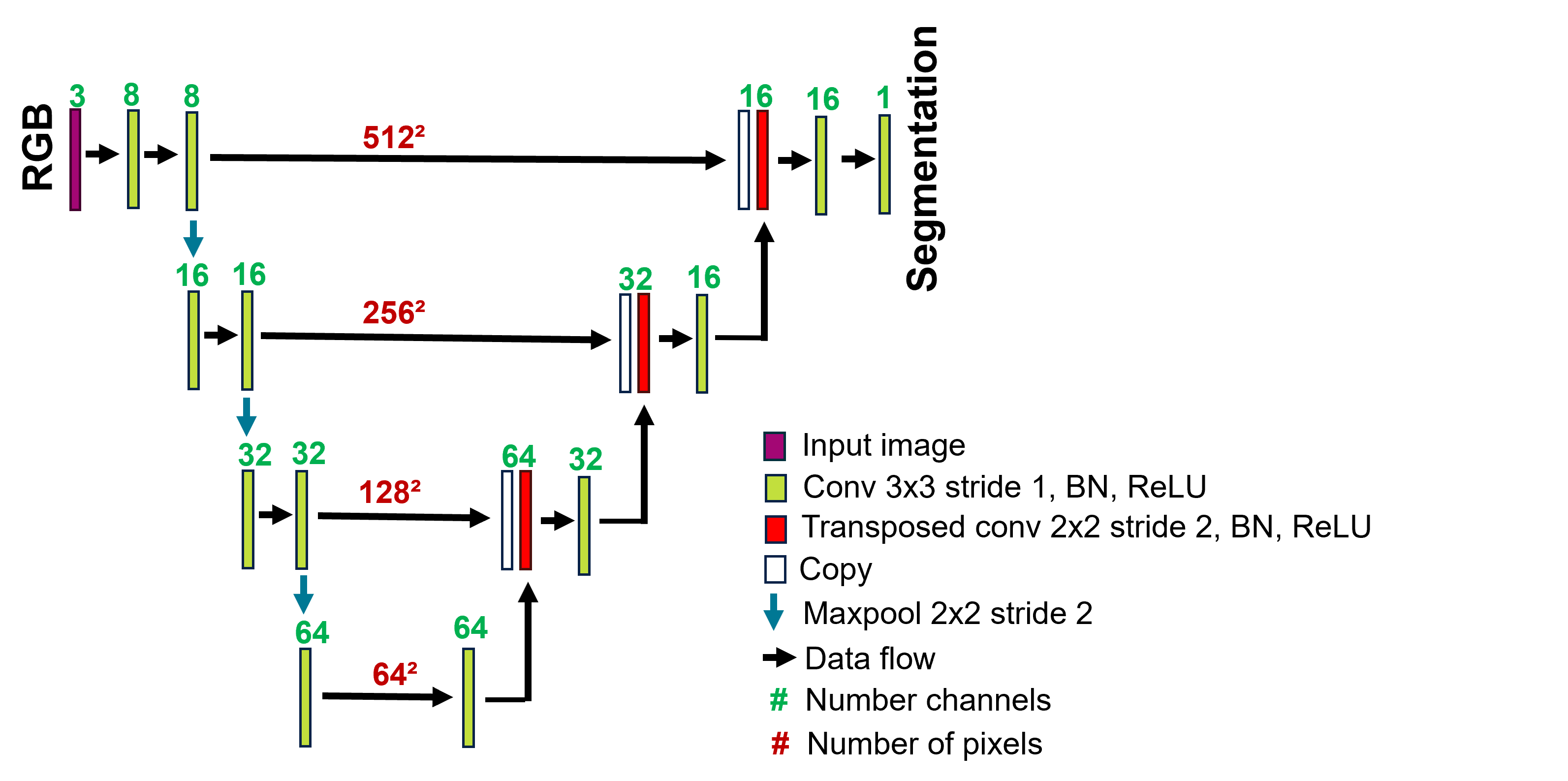}
    \caption{Baseline: \textit{Small U-Net model} architecture with a total of 110’689 parameters.
}
    \label{fig:baseline_architecture}
\end{figure*}

We have implemented a simpler baseline model in addition to our over-parameterized proposed model shown in Figure~\ref{fig:Res-U-Net}. The architecture of \textit{Small U-Net} is illustrated in Figure~\ref{fig:baseline_architecture} and is inspired by the original U-Net architecture \citep{U_Net}. The model includes three max-pooling-based downsampling stages with skip connections that copy the intermediate feature maps to the upsampling part, which utilizes transposed convolutions \citep{deconvolution}. A batch normalization layer \citep{batchnorm} and a ReLU activation function follow each convolution operation.

\subsection{Training details} 
\label{Appendix:segmentation_trainig_details}
All binary segmentation models were trained using the Adam optimizer \citep{adam}. The training protocol incorporates a learning rate warm-up phase of 100 iterations with batch size 32, during which the learning rate linearly increases from 0 to the base learning rate, followed by a cosine annealing schedule for the remaining steps. Gradient clipping was applied to ensure gradients did not surpass a maximum norm of 1 during training. The Dice Loss function was employed to address the class imbalance between road and non-road pixels \citep{DiceLoss}. Min-max normalization was applied to the image data by scaling pixel values from their original range of 0 to 255 to a normalized range of 0 to 1. This preprocessing step ensures uniformity across the dataset, aiding in improved performance and faster convergence of the deep learning model \citep{InputNorm}. The training spanned 50 epochs. We applied early stopping regularization to all models \citep{EarlyStopping}: During training, we evaluated the models on the validation set after each epoch, monitoring the validation scores based on the Intersection over Union (\textit{IoU}) criterion. The final model weights were selected based on the highest validation score achieved during training. For \textit{Attention ResU-Net} models, a dropout rate of 0.3 was utilized during training. The training process was executed on a 11GB GTX 1080 Ti GPU using \texttt{torchvision 0.17.2}.

\subsection{Model pre-training}
\label{Appendix:Model_pretraining}
As described in the main paper, we experimented with transfer learning by initializing our model with pre-trained weights before training on the \textit{Siegfried Map}. Initially, we loaded weights pre-trained on \textit{ImageNet} using PyTorch into the encoder part of our \textit{Attention ResU-Net} model \citep{Deng2009ImageNet}, which is based on the ResNet-18 classification model \citep{He2015DeepRL}. Subsequently, we trained the entire \textit{Attention ResU-Net} model on the \textit{Swiss Map} dataset. In this stage, we followed the procedure detailed in Appendix~\ref{Appendix:segmentation_trainig_details}, except that we used a base learning rate of 0.0005 with 3000 warm-up iterations and trained for 20 epochs using 19 \textit{Swiss Map} sheets. The model was evaluated on a validation set consisting of 5120 patches with a resolution of 500x500 pixels. The scores can be found in Table~\ref{tab:appendix_val_swissmap}.

\begin{table*}[h]
    \centering
    \tiny
    \resizebox{1.0\textwidth}{!}{
    \begin{tabular}{lccccccc}
    \toprule 
       Model & Pre-trained & \textit{Accuracy} & \textit{F1} & \textit{Precision} & \textit{Recall} & \textit{IoU} \\
        \midrule
        \textit{Attention ResU-Net} & \textit{ImageNet} & 98.55\%  & 96.63\% & 96.44\% & 96.83\% & 88.84\% \\ 
    \bottomrule
    \end{tabular}}
    \raggedright
    \caption{Performance results of the model on the \textit{Swiss Map} validation set during the pre-training phase.}
    \label{tab:appendix_val_swissmap}
\end{table*}

\subsection{Hyperparameter tuning}
\label{Appendix:segmentation_hyperparameter_tuning}
We performed hyperparameter tuning to determine the best base learning rate for each type of model we studied. Table~\ref{tab:segmentation_hyperparameter_tuning} displays the $Accuracy$ score and $IoU$ for all the learning rates we evaluated on the validation set. Additionally, Table~\ref{tab:segmentation_early_stopping} indicates the selected epoch for model selection based on the $IoU$ score on the validation set.

\begin{table*}[ht] 
    \centering
    \tiny
    \resizebox{1.0\textwidth}{!}{
    \begin{tabular}{lcccccccc}
    \toprule 
       Model & Pre-trained  & \multicolumn{2}{c}{Learning rate 0.1} & \multicolumn{2}{c}{Learning rate 0.01} & \multicolumn{2}{c}{Learning rate 0.001} \\
        \cmidrule(lr){3-4} \cmidrule(lr){5-6} \cmidrule(lr){7-8} 
        &&$Accuracy$&$IoU$
        &$Accuracy$&$IoU$ 
        &$Accuracy$&$IoU$\\ 
        \midrule
        \textit{Small U-Net} & No &  \textbf{97.76\%} & \textbf{82.27\%} &97.73\%& 81.95\% & 97.19\% & 78.76\%  \\
        \textit{Attention ResU-Net}  & No  & 97.72\% & 82.21\% & 97.74\% & 82.44\% & \textbf{97.78\%} & \textbf{82.66\%} \\
        \textit{Attention ResU-Net} & \textit{ImageNet}  & 97.68\% & 81.77\% & \textbf{97.91\%} & \textbf{83.43\%} & 93.14\% & 59.78\% \\
        \textit{Attention ResU-Net} & \textit{ImageNet} + \textit{Swiss Map}  & 97.77\% & 82.58\% & 98.08\% & 84.75\% & \textbf{98.11\%} & \textbf{84.95\%} \\
    \bottomrule  
    \end{tabular}}
    \caption{Results of evaluating different learning rates during hyperparameter tuning for each model on the validation set, with the best scores based on the selected learning rate for each model highlighted.}
    \label{tab:segmentation_hyperparameter_tuning}
\end{table*}

\begin{table*}[ht]
    \centering
    \tiny
    \resizebox{1.0\textwidth}{!}{
    \begin{tabular}{lcccc}
    \toprule 
       Model & Pre-trained  & \multicolumn{1}{c}{Learning rate 0.1} & \multicolumn{1}{c}{Learning rate 0.01} & \multicolumn{1}{c}{Learning rate 0.001} \\
        \cmidrule(lr){3-5} 
        &&Best epoch
        &Best epoch
        &Best epoch\\
        \midrule
        \textit{Small U-Net} & No &  47/50 & 41/50 & 30/50  \\
        \textit{Attention ResU-Net}  & No  & 41/50 & 43/50 & 35/50  \\
        \textit{Attention ResU-Net} & \textit{ImageNet}  & 31/50 & 38/50 & 14/50  \\
        \textit{Attention ResU-Net} & \textit{ImageNet} + \textit{Swiss Map}  & 40/50 & 45/50 & 37/50 \\
    \bottomrule
    \end{tabular}}
    \caption{Best-performing epoch selected for early stopping regularization for each model, along with the evaluated hyperparameter. The $IoU$ score on the validation set served as the early stopping metric, with the model from the best-performing epoch being selected.}
    \label{tab:segmentation_early_stopping}
\end{table*}

\subsection{Effect of data augmentation}
\label{Appendix:data_augmentation}
Given the limited size of the training data in this study, data augmentation can potentially be used to artificially increase the dataset size. However, if the augmented data significantly deviates from the original data distribution, it may negatively impact generalization performance \citep{MUMUNI2022100258}. Therefore, we analyzed the effect of data augmentation on validation set performance. Specifically, we compared no data augmentation with horizontal/vertical flipping combined with either random continuous rotations between 0-360 degrees or discrete rotations of 0, 90, 180, or 270 degrees. For these experiments, we used the baseline \textit{Small U-Net} model described in Appendix~\ref{Appendix:segmentation_trainig_details} trained with a learning rate of 0.1.

Table~\ref{tab:data_augmentation_study} shows the comparison results. It's clear that continuous rotation reduces the model's generalization ability. This is due to the fact that the coordinate grid lines are consistently horizontal and vertical, leading to a scenario where the model cannot learn to distinguish these lines from roads with similar symbology. On the other hand, discrete rotation enhances performance by preserving the inherent structure of the coordinate grid while artificially increasing the training data size.

\begin{table*}[ht]
    \centering
    \tiny
    \resizebox{1.0\textwidth}{!}{
    \begin{tabular}{lcccccc}
    \toprule 
       Model & Augmentation & Best epoch & \textit{Accuracy} & \textit{IoU}\\

        \midrule  
        \textit{Small U-Net} & No &  32/50 & 97.58\% & 80.19\% \\
        \textit{Small U-Net} & Horizontal/vertical flip (p=0.5), rotation (0°, 90°, 180°, 270°) & 47/50 & \textbf{97.76\%} & \textbf{82.27\%}\\
        \textit{Small U-Net} & Horizontal/vertical flip (p=0.5), rotation (0°-360°) & 43/50 & 97.18\% & 78.91\%\\
    \bottomrule
    \end{tabular}}
    \caption{The effect of data augmentation on the validation set performance. All models were trained with early stopping based on IoU and a learning rate of 0.1.}
    \label{tab:data_augmentation_study}
\end{table*}

\subsection{Model selection}
\label{Appendix:model_selection}
The model selection was based on the performance of the validation set. Table~\ref{tab:appendix_validation} presents the results for different models on the validation set. It is clear that our proposed method, \textit{Attention ResU-Net}, performs better than our baseline, \textit{Small U-Net}. Furthermore, the results show that initializing the encoder part of \textit{Attention ResU-Net} with \textit{ImageNet} pre-trained weights leads to slightly better performance than random initialization. However, the best-performing model was the \textit{Attention ResU-Net}, which was first initialized with an \textit{ImageNet} pre-trained encoder before being pre-trained on \textit{Swiss Map}. Then, the model was fine-tuned on the \textit{Siegfried Map}. This model was finaly selected for our classification framework.

\begin{table*}[h]
    \centering
    \scriptsize
    \resizebox{1.0\textwidth}{!}{
    \begin{tabular}{lcccccc}
    \toprule 
       Model & Pre-trained & \textit{Accuracy} & \textit{F1} & \textit{Precision} & \textit{Recall} & \textit{IoU}\\ 
       
        \midrule
        \textit{Small U-Net} & No & 97.76\% & 94.50\% & 94.23\% & 94.78\% & 82.27\%  \\ 
        \textit{Attention ResU-Net}  & No & 97.78\% & 94.63\% & 93.92\% & 95.37\% & 82.66\% \\ 
        \textit{Attention ResU-Net} & \textit{ImageNet} & 97.91\% & 94.89\% & 94.46\% & 95.34\% & 83.43\% \\ 
        \textit{\underline{Attention ResU-Net}} & \textit{ImageNet} + \textit{Swiss Map}& \textbf{98.11\%} & \textbf{95.40\%} & \textbf{94.89\%} & \textbf{95.92}\% & \textbf{84.95\%} \\ 
    \bottomrule
    \end{tabular}}
    \caption{Results on validation set for each model using best performing hyperparameters. Best scores are highlighted, final selected model is underlined. }
    \label{tab:appendix_validation}
\end{table*}

\subsection{Performance on test data}
\label{Appendix:generlalization_performance}
We evaluated all models on the test dataset as an additional study for research purposes without conducting model selection for our framework. Table~\ref{tab:appendix_test} presents the results for all the implemented models on the test data. The selected model \textit{Attention ResU-Net} pre-trained on \textit{Swiss Map} achieves the best scores overall on the test data.

\begin{table*}[h]
    \centering
    \scriptsize
    \resizebox{1.0\textwidth}{!}{
    \begin{tabular}{lccccccc}
    \toprule 
       Model & Pre-trained & \textit{Accuracy} & \textit{F1} & \textit{Precision} & \textit{Recall} & \textit{IoU} \\ 
       
        \midrule
        \textit{Small U-Net} & No & 97.76\% & 92.85\% & 90.26\% & 95.88\% & 76.86\% \\ 
        \textit{Attention ResU-Net}  & No & 98.07\% & 93.70\% & 91.82\% & 95.81\% & 79.30\% \\ 
        \textit{Attention ResU-Net} & \textit{ImageNet} & 98.08\% & 93.47\% & 93.21\% & 93.73\% & 78.55\% \\
        \textit{\underline{Attention ResU-Net}} & \textit{ImageNet} + \textit{Swiss Map} & \textbf{98.37\%}  & \textbf{94.64\%} & \textbf{93.05\%} & \textbf{96.39\%} & \textbf{82.10\%} \\ 
    \bottomrule
    \end{tabular}}
    \raggedright
    \caption{Results of all models on the test set. Selected model based on the validation set is underlined, best scores are highlighted.}
    \label{tab:appendix_test}
\end{table*}

\section{Road classification model}
\label{Appendix:classification_model}
This Section provides further details about the road classification model, including training details and additional results. The metrics used in this chapter to assess the models' performance are specified in detail in Appendix \ref{Appendix:evaluation_metrics}.

\subsection{Training details}
\label{Appendix:classification_model_training_details}
The classification model was trained using the Adam optimizer with a weight decay of 0.00001 for regularization \citep{adam}. A constant learning rate of 0.0005 and a batch size of 16 were used. Gradient clipping was applied to prevent gradients from exceeding a maximum norm of 1 during training. The image data was normalized using min-max normalization, which scaled pixel values from the original range of 0 to 255 to a normalized range of 0 to 1. Additionally, unlike the binary segmentation model, no dropout or data augmentation was used. This decision was made because the model is fine-tuned for only two epochs, during which data augmentation could cause a larger distribution shift rather than effectively increasing the training data size. Moreover, dropout hinders the model's convergence and results in unstable performance after only two epochs of training. Label smoothing with an epsilon parameter of 0.05 was applied to the cross-entropy loss function to enhance calibration and regularization \citep{label_smoothing}. The model weights were initialized using the final weights of a binary segmentation model, except for the last layer, which was initialized randomly. The final model is an ensemble of 30 models, each trained with different initializations of the last layer and different orders of images during training. During inference, the ensemble members' class likelihoods are averaged to receive the final predictions. The training process was carried out on an 11GB GTX 1080 Ti GPU using \texttt{torchvision 0.17.2}.

\subsection{Optimizing training epochs}
\label{Appendix:classification_model_epochs}
\begin{figure*}[h]
    \subfloat[$Accuracy$ ($\uparrow$)]{{\includegraphics[width=0.33\textwidth]{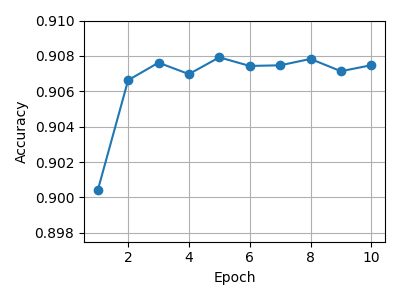}\label{fig:epoch_tuning_accuracy}}}%
    \subfloat[$F1~Score$ ($\uparrow$)]{{\includegraphics[width=0.33\textwidth]{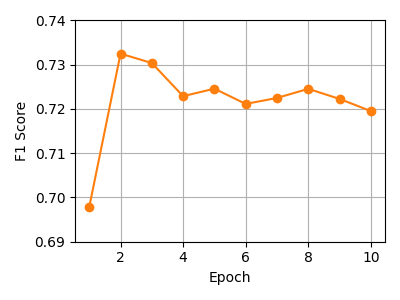}\label{fig:epoch_tuning_f1}}}%
    \subfloat[$IoU$ ($\uparrow$)]{{\includegraphics[width=0.33\textwidth]{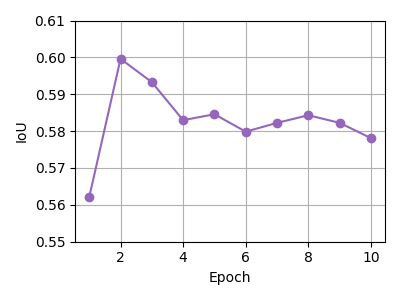}\label{fig:epoch_tuning_IoU}}}%
    \qquad
    \subfloat[$Precision$ ($\uparrow$)]{{\includegraphics[width=0.33\textwidth]{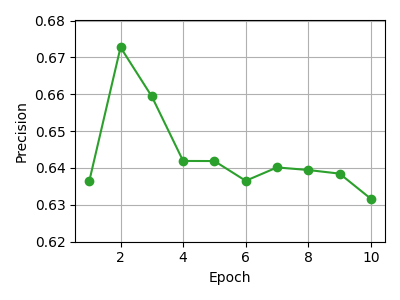} \label{fig:epoch_tuning_precision}}}%
    \subfloat[$Recall$ ($\uparrow$)]{{\includegraphics[width=0.33\textwidth]{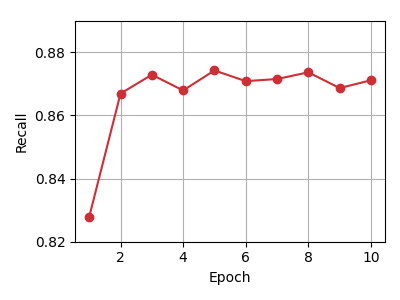}\label{fig:epoch_tuning_recall}}}%
    \subfloat[$Brier~Score$ ($\downarrow$)]{{\includegraphics[width=0.33\textwidth]{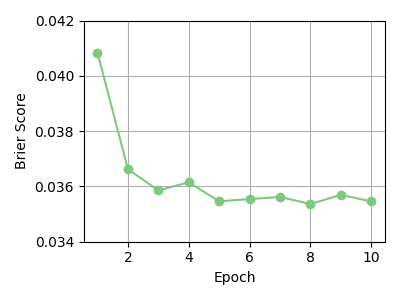} \label{fig:epoch_tuning_brier}}}%
    \caption{Evaluation metrics depend on the number of trained epochs, with results based on synthetic validation data.}
    \label{fig:epoch_tuinng}
\end{figure*}

Since we initialize the classification model with the binary road segmentation model weights, only minimal finetuning is needed since the binary segmenation model has already implicitly learned to detect various types of roads. This is evident as shown in Figure~\ref{fig:epoch_tuinng}: $Accuracy$, $Recall$, and $Brier~Score$ on the synthetic validation set have only minimal improvements after two training epochs, while $F1~Score$, $IoU$, and $Precision$ even decrease after two epochs. Therefore, we chose a training procedure involving two epochs of finetuning for our framework. With 30 members in our ensemble, we only require 60 training epochs in total, making our approach computationally efficient.

\subsection{Effect of distribution shift: More results}
\label{Appendix:classification_model_ensemble_size}
\begin{figure*}[h]
    \subfloat[$Accuracy$ ($\uparrow$)]{{\includegraphics[width=0.33\textwidth]{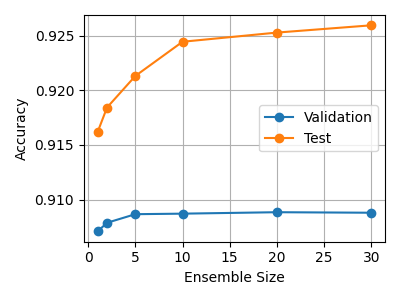}\label{fig:ensemble_size_accuracy_appendix}}}%
    \subfloat[$F1~Score$ ($\uparrow$)]{{\includegraphics[width=0.33\textwidth]{Images/F1_ensemble_size.png}\label{fig:ensemble_size_F1_appendix}}}%
    \subfloat[$IoU$ ($\uparrow$)]{{\includegraphics[width=0.33\textwidth]{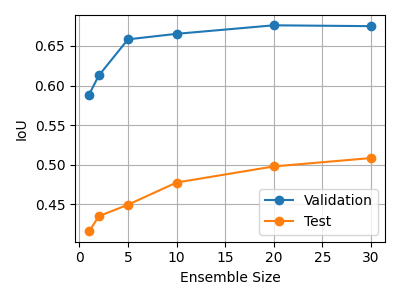}\label{fig:ensemble_size_IoU_appendix}}}%
    \qquad
    \subfloat[$Precision$ ($\uparrow$)]{{\includegraphics[width=0.33\textwidth]{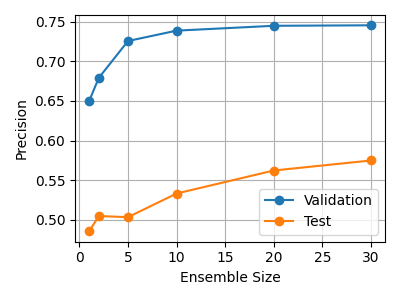} \label{fig:ensemble_size_precision_appendix}}}%
    \subfloat[$Recall$ ($\uparrow$)]{{\includegraphics[width=0.33\textwidth]{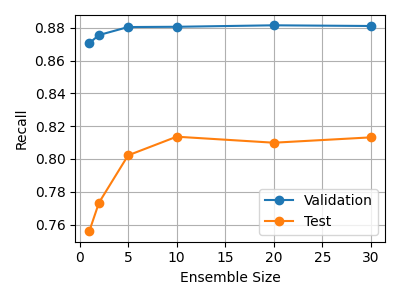}\label{fig:ensemble_size_recall_appendix}}}%
    \subfloat[$Brier~Score$ ($\downarrow$)]{{\includegraphics[width=0.33\textwidth]{Images/Brier_ensemble_size.png} \label{fig:ensemble_size_brier_appendix}}}%
    \caption{Evaluation metrics dependent on the ensemble size for synthetic validation and original test data.}
    \label{fig:ensemble_size_appendix}
\end{figure*}

This Section presents additional results on the effect of distribution shift on the predictive performance and calibration of the classification model discussed in Section~\ref{sec:effect_dist_shift}. Figure~\ref{fig:ensemble_size_appendix} displays evaluation metrics, including $Accuracy$, $F1~Score$, $IoU$, $Recall$, $Precision$, and $Brier~Score$, for varying ensemble sizes.

The problematic distribution shift is evident from the discrepancy between the model's performance on the synthetic validation set and the original test set. Ensembling improves performance across all metrics. While the $Accuracy$ and $Brier~Score$ are superior on the test set than on the validation set, this pattern should be interpreted cautiously since the metrics include also the majority class of "no road" pixels. This class is predicted by the hard masking mechanism, meaning that it is not affected by the distribution shift.

\section{Evaluation metrics}
\label{Appendix:evaluation_metrics}
This Section presents the definitions of the evaluation metrics we used to assess the performance of our segmentation models.

\subsection{\textit{Accuracy}}
\begin{equation}
Accuracy = \frac{1}{N} \sum_{i=1}^{N} \mathbf{1}(\hat{y}_i = y_i).
\end{equation}
Here $y_i$ represents the true label of pixel $i$, $\hat{y}_i$ represents the predicted label of pixel $i$, and $N$ denotes the total number of evaluated pixels.

\subsection{\textit{F1 Score}}
We define the $F1~Score$ in this paper as the macro variant:
\begin{equation}
    (Macro)~F1=\frac{1}{C}\sum_{j=1}^{C}\frac{2p_jr_j}{p_j+r_j},
\end{equation}
where $r_j $ is the Recall for class $ j $, given by $r_j = \frac{TP}{TP + FN} $, and $p_j $ is the Precision for class $ j $, defined as $p_j = \frac{TP}{TP + FP} $. In this context, $ C $ represents the total number of classes. The terms $TP$, $FP$, and $FN$ stand for true positives, false positives, and false negatives, respectively.

\subsection{\textit{Intersection over Union (IoU)}}
\begin{equation}
    IoU(A,B) = \frac{\left | A \cap B \right |}{\left |A\cup B\right |},
\end{equation}
where $A$ represents the ground truth set and $B$ represents the prediction set. This metric ranges from 0 to 1, where 0 indicates no overlap, and 1 indicates a perfect match.

\subsection{\textit{Precision}}
We define the $Precision$ score in this paper as the macro variant:
\begin{equation}
    (Macro)~Precision = \frac{\sum_{j=1}^{C} Precision_j}{|C|},
\end{equation}

\begin{equation}
    Precision_j = \frac{TP_j}{TP_j + FP_j},
\end{equation}
where $Macro~Precision$ is the average $Precision$ across all classes, $C$ is the total number of classes, and $Precision_j$ is the $Precision$ for class $j$. Here, $TP_j$ represents the number of correctly predicted positive samples for class $j$, and $FP_j$ represents the number of incorrectly predicted positive samples for class $j$.

\subsection{\textit{Recall}}
We define the $Recall$ score in this paper as the macro variant:
\begin{equation}
    (Macro)~Recall = \frac{\sum_{j=1}^{C} Recall_j}{|C|},
\end{equation}

\begin{equation}
    Recall_j = \frac{TP_j}{TP_j + FN_j},
\end{equation}
where $Macro~Recall$ is the average $Recall$ across all classes, $C$ is the total number of classes, and $Recall_j$ is the $Recall$ for class $j$. Here, $TP_j$ represents the number of correctly predicted positive samples for class $j$, and $FN_j$ represents the number of incorrectly predicted negative samples for class $j$.

\subsection{\textit{Brier Score}}

The $Brier~Score$ is a commonly used metric for evaluating the calibration of neural networks \citep{Brier1950}. It is a variant of the mean squared error applied to predicted probabilities. The calculation of the $Brier~Score$ is as follows:

\begin{equation}
    Brier~Score = \frac{1}{N}\sum_{i=1}^{N}\sum_{j=1}^C(\hat{p}_{i,j} - y_{i,j})^2.
\end{equation}

Here, $N$ represents the number of evaluated pixels, $C$ denotes the number of classes, $y_{i,j}$ is 1 if the true label of pixel $i$ is $j$ and 0 otherwise, and $\hat{p}_{i,j}$ is the predicted probability of pixel $i$ belonging to class $j$.

\end{document}